\pdfoutput=1

\documentclass[11pt]{article}

\usepackage{acl}

\usepackage{times}
\usepackage{latexsym}

\usepackage[T1]{fontenc}

\usepackage[utf8]{inputenc}

\usepackage{microtype}

\usepackage{times}
\usepackage{latexsym}
\usepackage{enumitem}
\usepackage{multirow}
\usepackage{array}
\usepackage{amsfonts}
\usepackage{subcaption}
\usepackage{makecell}
\usepackage{soul}
\usepackage{xcolor,colortbl}
\usepackage{xspace}

\usepackage{arydshln}
\usepackage{textcomp}
\usepackage{stfloats}
\usepackage{url}
\usepackage{verbatim}
\usepackage{graphicx}
\usepackage{algorithm}
\usepackage{algorithmic}
\usepackage{amssymb}
\usepackage{helvet}  
\usepackage{courier}  
\usepackage{newfloat}
\usepackage{listings}
\usepackage{multirow}
\usepackage{amsmath}
\usepackage{booktabs}
\usepackage{pifont}
\usepackage{color}

\newcommand{{\model}}{KICT}

\title{Knowledgeable In-Context Tuning:\\ Exploring and Exploiting Factual Knowledge for In-Context Learning}

\author{Jianing Wang$^1$\thanks{\ \ \ J. Wang and C. Wang contributed equally to this work.}, Chengyu Wang$^{2*}$, Chuanqi Tan$^2$, Jun Huang$^2$, Ming Gao$^{1,3}$\thanks{\ \ \ Corresponding author.}\\
  $^1$ School of Data Science and Engineering, East China Normal University, Shanghai, China\\
  $^2$ Alibaba Group, Hangzhou, China\\
  $^{3}$ KLATASDS-MOE, School of Statistics, East China Normal University, Shanghai, China \\
  \texttt{lygwjn@gmail.com,\{chengyu.wcy,chuanqi.tcq\}@alibaba-inc.com}\\
  \texttt{huangjun.hj@alibaba-inc.com,mgao@dase.ecnu.edu.cn}\\}

\begin{document}
\maketitle
\begin{abstract}

Large language models (LLMs) enable in-context learning (ICL) by conditioning on a few labeled training examples as a text-based prompt, eliminating the need for parameter updates and achieving competitive performance.
In this paper, we demonstrate that \emph{factual knowledge} is imperative for the performance of ICL in three core facets: the inherent knowledge learned in LLMs, the factual knowledge derived from the selected in-context examples, and the knowledge biases in LLMs for output generation.
To unleash the power of LLMs in few-shot learning scenarios, we introduce a novel \textbf{K}nowledgeable \textbf{I}n-\textbf{C}ontext \textbf{T}uning (\textbf{KICT}) framework to further improve the performance of ICL:
1) injecting knowledge into LLMs during continual self-supervised pre-training, 2) judiciously selecting the examples for ICL with high knowledge relevance, and 3) calibrating the prediction results based on prior knowledge.
We evaluate the proposed approaches on autoregressive models (e.g., GPT-style LLMs) over multiple text classification and question-answering tasks.  Experimental results demonstrate that \textbf{KICT} substantially outperforms strong baselines and improves by more than 13\% and 7\% on text classification and question-answering tasks, respectively~\footnote{The code and datasets are released in HugNLP~\cite{Wang2023HugNLP}:~\url{https://github.com/HugAILab/HugNLP}.}.

\end{abstract}

\section{Introduction}
\begin{figure}
\centering
\includegraphics[width=\linewidth]{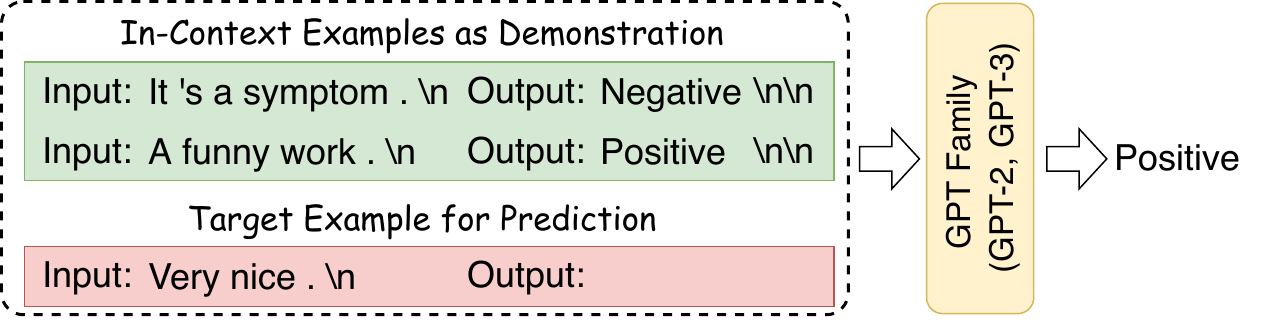}
\caption{An example of in-context learning (ICL).}
\vspace{-.5em}
\label{fig:example}
\end{figure}

Large language models (LLMs) have become an imperative infrastructure in the natural language processing (NLP) community~\cite{DBLP:journals/corr/abs-2303-18223}. To enable pre-trained LLMs to perform well without any parameter updates, in-context learning (ICL) has emerged as one of the flourishing research topics in many few-shot NLP tasks. It aims to generate predictions for target examples by conditioning on a few labeled samples~\cite{Brown2020Language}.
As shown in Figure~\ref{fig:example}, the key component of ICL is the text-based prompt (containing labeled examples) that functions as the demonstration.

Previous works have explored multiple aspects that affect the performance of ICL~\cite{Dong2023A}, such as input-output mapping~\cite{Min2022Rethinking, Kim2022Ground}, extensive data resources~\cite{Mishra2022Cross, Chen2022Meta, Min2022MetaICL}, prediction calibration~\cite{Zhao2021Calibrate}, and self-improvment~\cite{Chen2023SelfICL, Lyu2023ZICL}. 
\citet{Liu2022What, Yao2022Fantastically} have investigated others, such as prompt format (e.g., ``Input:'', ``Output:''), the selection of labeled data, and example permutation. 
\citet{Wang2023HugNLP, Wu2023OpenICL} have developed toolkits for LLMs to reason with ICL prompts.
In addition, to better elicit the LLM to reason on complex tasks, chain-of-thought (CoT) has been introduced to extend the ICL with multiple rationales to express the thinking process~\cite{Wei2022Chain, Dhuliawala2023Chain, Wang2023Self, Wang2023Boosting, Zhao2023Verify, Zhang2023Automatic, Liang2023Prompting}.
However, these works pay little attention to the influence of \emph{factual knowledge} in ICL, which is a non-negligible factor in NLP~\cite{Hu2022Knowledgeable}.

\begin{figure*}
\centering
\begin{tabular}{cc}
\begin{minipage}[t]{0.475\linewidth}
    \includegraphics[width = \linewidth]{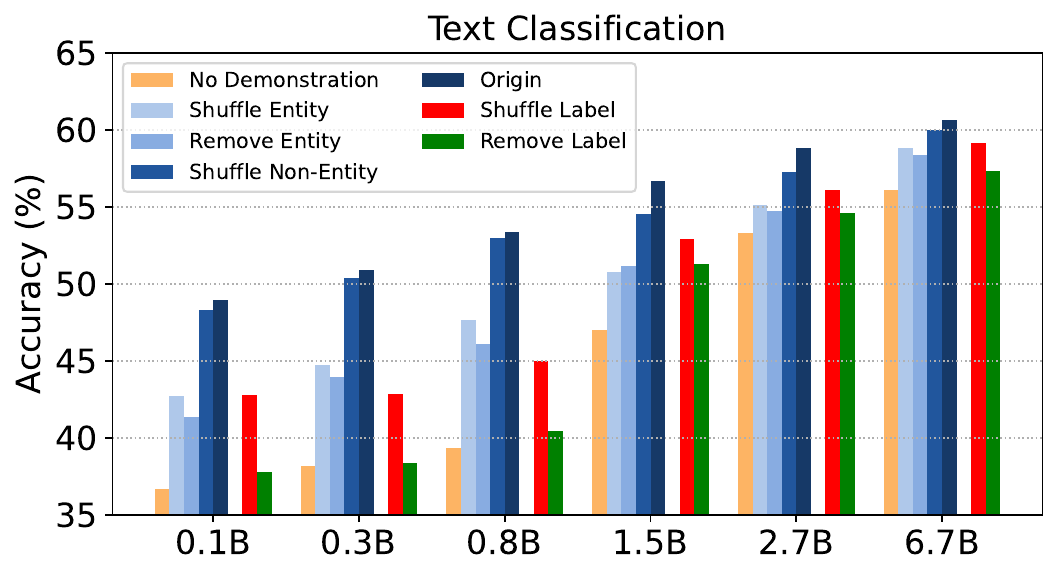}
\end{minipage}
\begin{minipage}[t]{0.475\linewidth}
    \includegraphics[width = \linewidth]{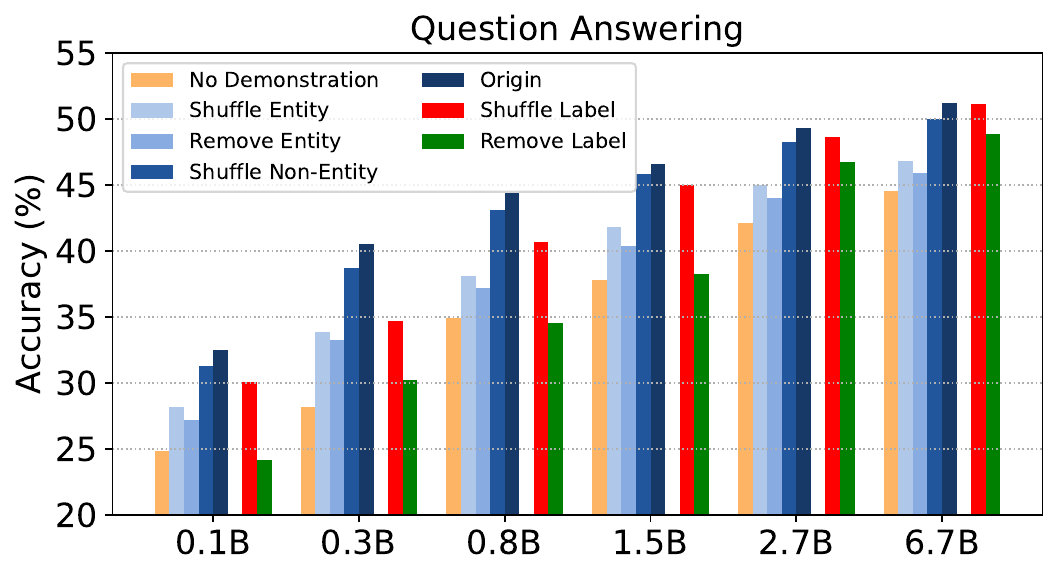}
\end{minipage}
\end{tabular}
\caption{Results of different scales of GPT-2 and OPT models over 8 text classification tasks and 4 question answering tasks in various component destruction settings. For each target example, we have $K=8$ labeled samples as the demonstration. Results indicate that factual knowledge is crucial to the performance of ICL.}
\label{fig:preexp1}
\end{figure*}

To this end, we explore the effectiveness of ICL from the perspective of \emph{factual knowledge}.
As seen in Figure~\ref{fig:preexp1}, when entities and labels in text-based prompts are randomly replaced or removed, the average accuracy decreases significantly, indicating that performance degradation is universal across different model scales.
Further analysis reveals that: 1) more intrinsic factual knowledge acquired during the pre-training stage is typically beneficial for LLMs to improve effectiveness;
2) The factual knowledge (e.g., entities and labels) derived from selected in-context examples is crucial for the performance of ICL;
3) LLMs tend to generate common words that may have high frequencies in the training corpora, resulting in biased predictions.

After analyzing these knowledge facets, a natural question arises: \emph{How can we fully employ factual knowledge to further improve the performance of ICL?}
To achieve this goal, we focus on causal autoregressive LLMs (e.g., GPT-2~\cite{radford2019language} and OPT~\cite{Zhang2022OPT}) and present a novel \textbf{K}nowledgeable \textbf{I}n-\textbf{C}ontext \textbf{T}uning (\textbf{KICT}) framework, which involves knowledgeable guidance in \emph{pre-training}, \emph{prompting}, and \emph{prediction} of these models.
Specifically, to endow LLMs with enhanced text generation abilities by better leveraging inherent knowledge, we introduce several knowledgeable self-supervised tasks during the \emph{pre-training} stage to inject knowledge into LLMs. 
For text-based \emph{prompting}, we propose a knowledgeable example retrieval algorithm to judiciously select in-context examples that have relevant knowledge to the target example. 
Finally, during \emph{prediction}, we utilize the knowledge-wise priors of label words from an underlying knowledge base (KB) to calibrate the prediction distributions generated by LLMs.
Each of the proposed techniques is plug-and-play and can be freely combined, facilitating users to exploit knowledge for improving ICL.

To evaluate the effectiveness of the \textbf{KICT} framework, we employ LLMs (e.g., GPT-style models) to conduct extensive experiments over multiple text classification and question-answering tasks.
Results demonstrate that each proposed procedure achieves substantial improvements.

To sum up, we make the following main contributions:
\begin{itemize}
\item We study three knowledge facets for ICL that are imperative for LLMs in few-shot learning, i.e., inherent knowledge in LLMs, relevant knowledge in the text-based prompt, and knowledge bias.

\item We present a novel knowledgeable in-context tuning framework for better incorporating knowledge through the process of pre-training, prompting, and predicting.

\item Extensive experiment results show that our approach attains more impressive performance over classification and QA tasks.
\end{itemize}

\section{Impact of Knowledge on ICL}
\label{sec:preliminary}


In this section, we investigate whether \emph{factual knowledge} affects the performance of ICL.

\subsection{Preliminary Experimental Settings}
Following \citet{Min2022Rethinking} and \citet{Kim2022Ground}, we perform empirical experiments through component destruction.
Specifically, given a target example text $X^{tgt}$, we randomly select $K$ training samples $\widetilde{\mathcal{D}}=\{(X^{trn}_i, y^{trn}_i)\}_{i=1}^{K}$ to form a text-based prompt. We identify all entities in the prompt and then devise several destruction settings as follows: 1) \texttt{Shuffle Entity} involves randomly replacing all entities with others from the KB; 2) \texttt{Shuffle Non-Entity} entails replacing some non-entity words (e.g., ``It'', ``have'') with others from the vocabulary; 3) \texttt{Shuffle Label} consists of replacing all the golden labels with incorrect ones; 4) \texttt{Remove Entity} and \texttt{Remove Label} aim to remove all entities and labels from the prompt, respectively; 5) \texttt{No Demonstration} represents a typical zero-shot method where no labeled data is used~\cite{Min2022Rethinking}.

We employ various scales of GPT-2 (0.1B-1.5B) and OPT~\cite{Zhang2022OPT} (2.7B-6.7B) models to evaluate 8 text classification tasks and 4 question answering tasks.
\footnote{Due to resource constraints, we do not use larger models. Nevertheless, our findings are generally consistent across different model scales.}
By default, we randomly sample $K=8$ labeled samples for each task and conduct the experiments with 5 different random seeds.
Further details are presented in Appendix~\ref{appendix:preliminary}.
The findings are summarized below.

\subsection{Findings}


\noindent\textbf{The inherent knowledge in the LLM itself is beneficial for the performance of downstream tasks.}
As shown in Figure~\ref{fig:preexp1}, models can achieve remarkable few-shot performance with increased scale.
We hypothesize that larger models can learn more valuable semantics in the pre-training corpus, which contributes to this improvement.
To test this hypothesis, we perform zero-shot inference without any text-based prompts (i.e., \texttt{No Demonstration}), relying solely on the intrinsic knowledge acquired during pre-training to guide the predictions.
We observe that the performance gap between the 6.7B and 0.1B models is about 20\% on both text classification and question-answering tasks.
This observation supports the idea that the inherent knowledge learned during pre-training is critical~\cite{Yang2021A}.


\begin{figure}[t]
\centering
\begin{tabular}{ll}
\begin{minipage}[t]{0.49\linewidth}
    \includegraphics[width = 1\linewidth]{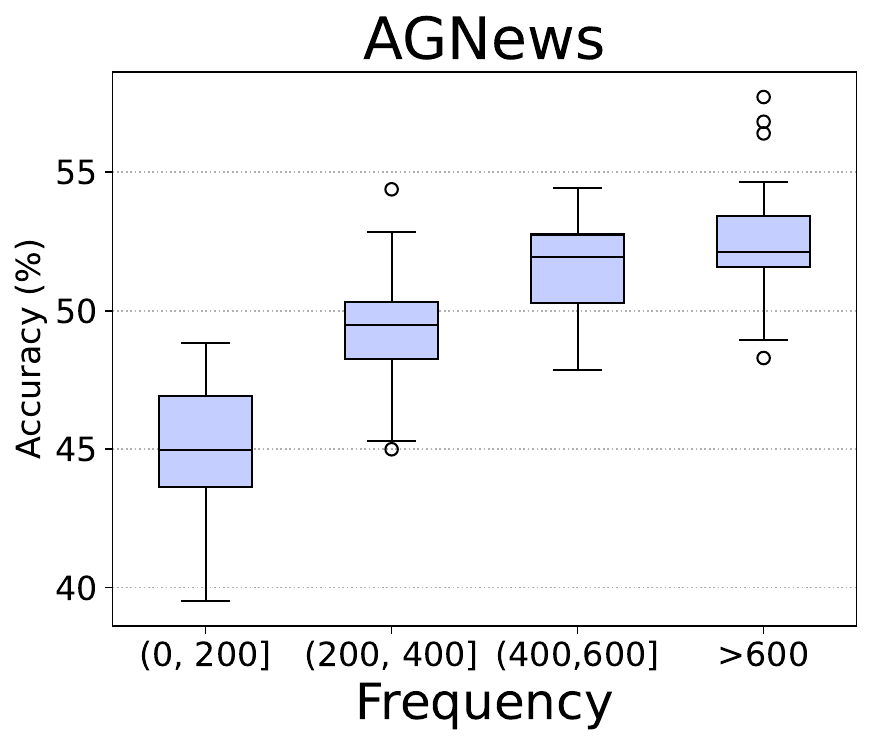}
\end{minipage}
\begin{minipage}[t]{0.49\linewidth}
    \includegraphics[width = 1\linewidth]{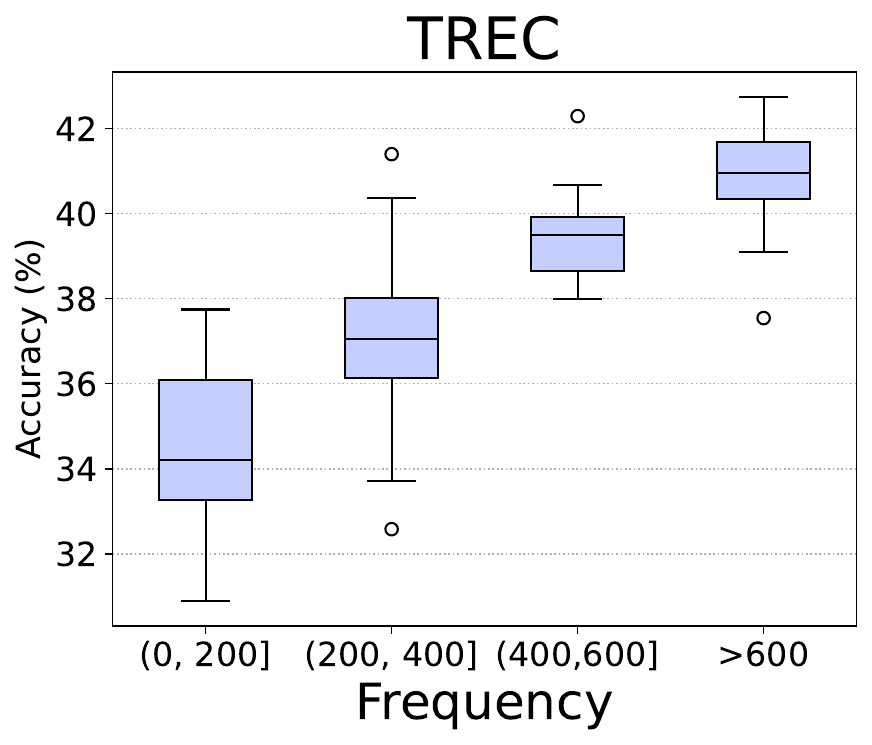}
\end{minipage}
\end{tabular}
\vspace{-.75em}
\caption{4-shot results of GPT-2 (urge) over AGNews and TREC. For each frequency region,  we sample top-5 label words for each category and report the accuracy for all label mapping permutations. }
\label{fig:preexp2}
\end{figure}

\begin{figure*}[t]
\centering
\includegraphics[width=\linewidth]{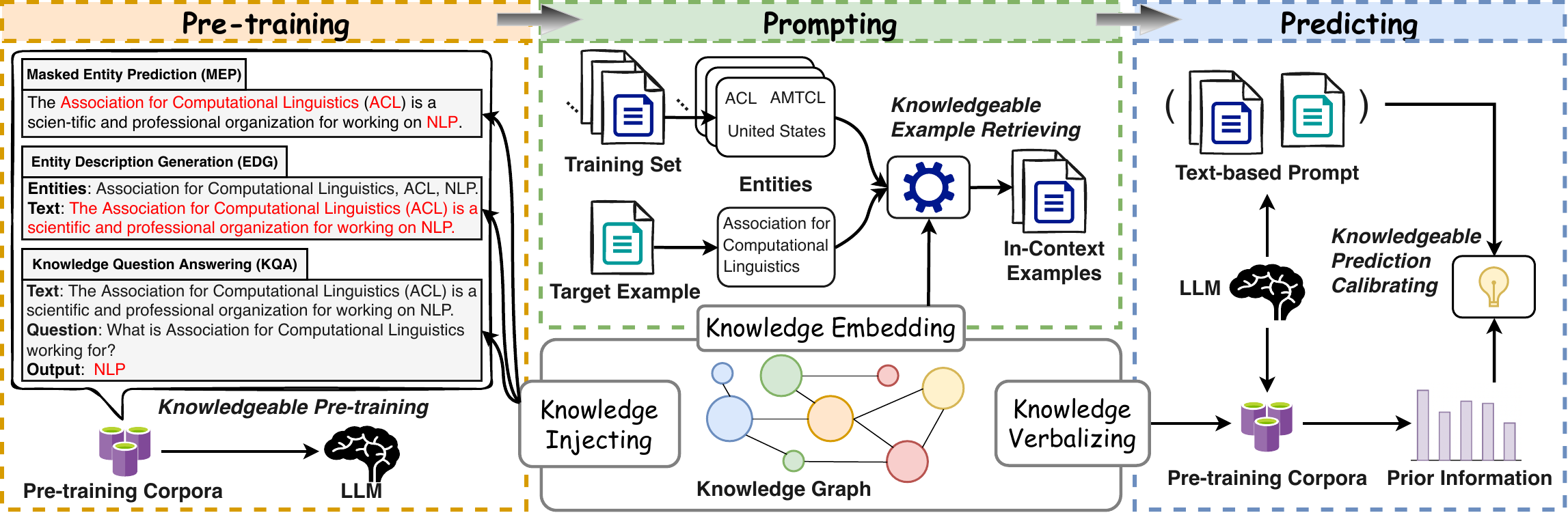}
\caption{The overview of the \textbf{KICT} framework. We introduce multiple plug-and-play knowledgeable techniques to enhance the utilization of knowledge for improving ICL performance.
\textbf{Left}: We propose three knowledge-aware self-supervised learning tasks that infuse factual knowledge into LLMs during pre-training.
\textbf{Middle}: We utilize entity-related information to select in-context examples that exhibit high knowledge relevance to the target example.
\textbf{Right}: For prediction, we derive prior information from large-scale corpora to calibrate the predictions.}
\label{fig:model}
\end{figure*}

\noindent\textbf{The factual knowledge in selected in-context examples is crucial for ICL.}
As shown in Figure~\ref{fig:preexp1}, the original setting (\texttt{Origin}) outperforms other configurations across all model scales.
We observe that altering non-entity words does not significantly reduce performance, whereas replacing or removing entities leads to a considerable decrease in average accuracy for both text classification and question-answering tasks.
This demonstrates that factual knowledge embedded in text-based prompts is a critical factor for LLMs to understand the task.
Furthermore, we find that labels are also essential for ICL, echoing similar observations presented in~\cite{Kim2022Ground}.
Differing from \citet{Min2022Rethinking}, we posit that labels can be regarded as a form of factual knowledge that guides the LLM to grasp semantics during inference.


\noindent\textbf{LLMs tend to generate common label words due to knowledge bias.}
To investigate whether predictions are biased, we select two knowledge-intensive tasks (i.e., AGNews~\cite{Zhang2015Character}, and TREC~\cite{Voorhees2000Building}). We first retrieve the top-5 predictions at the output position for each training example\footnote{The training set is larger than the testing set, thereby providing a more robust statistical representation.} and compute frequency statistics for each generated label word.
Subsequently, we select 4 labeled examples from the training set for each category.
From each frequency region, we randomly choose 2 label words and calculate the average accuracy across all label mapping permutations.\footnote{Considering AGNews as an example, which has 4 classes with 2 label words each, there are $2^4=16$ possible label mapping permutations.}
The results, as presented in Figure~\ref{fig:preexp2}, reveal that performance is highly contingent on label word frequency, suggesting that the frequency with which factual knowledge is learned by LLMs plays a critical role in prediction outcomes. Similar observations have been reported by \citet{Zhao2021Calibrate}.





\section{The Proposed \textbf{KICT} Framework}
The preliminary experiments demonstrate that \emph{factual knowledge} has a substantial effect on ICL. This suggests that we can exploit this knowledge to enhance performance across various processes in ICL, including \emph{pre-training}, \emph{prompting}, and \emph{prediction}.
To achieve this goal, we introduce the \textbf{KICT} framework, a novel \textbf{K}nowledgeable \textbf{I}n-\textbf{C}ontext \textbf{T}uning framework designed to better leverage knowledge and unleash the power of LLMs in answer generation. 
Within this framework, we introduce Knowledgeable Pre-Training (KPT) with three carefully designed self-supervised tasks to infuse LLMs with factual knowledge.
We then present a Knowledgeable Example Retrieval (KER) algorithm to judiciously select in-context examples that are relevant to the given knowledge.
Finally, we employ a Knowledgeable Prediction Calibration (KPC) technique to adjust the prediction distribution using prior information derived from a KB.
An overview of the framework is depicted in Figure~\ref{fig:model}.

\subsection{Knowledgeable Pre-Training}


This section describes three knowledge-aware self-supervised learning tasks designed to infuse factual knowledge into LLMs, namely, \emph{Masked Entity Prediction} (MEP), \emph{Entity Description Generation} (EDG), and \emph{Knowledgeable Question Answering} (KQA). Differing from \citet{Chen2022Improving}, we leverage an external KB to enrich the models' language generation abilities with respect to important entities.
The input consists of a training corpus $\{X\}$ and a KB $\mathcal{G}=(\mathcal{E}, \mathcal{R}, \mathcal{T})$,
where $\mathcal{E}$ denotes a set of entities, $\mathcal{R}$ a set of relations, and $\mathcal{T}$ a set of triples representing factual knowledge.

\noindent\textbf{Masked Entity Prediction (MEP).}
MEP requires the model to predict missing entities within a text, enhancing its capability to learn explicit knowledge. This task is akin to \emph{Masked Language Modeling} employed in BERT-style models~\cite{Devlin2019BERT, Liu2019RoBERTa}. 
Given a text composed of tokens $X=\{x_i\}$, we identify all entities $E_{X}=\{e|e\in\mathcal{G}, e\in X\}$ using an entity linking toolkit. Each entity $e=\{x_j|x_j\in X\}$, which may span multiple tokens, is either replaced with special tokens (e.g., ``\_'') or random tokens with equal probability. This process generates a modified text $\hat{X}=\{\hat{x}_i\}$.
A label mask vector $\mathcal{M}_{\hat{X}}$ is created to indicate training positions, where $\mathcal{M}_{\hat{X}_i}=\mathbb{I}(\hat{x}_i\in E_{X})$ and $\mathbb{I}(\cdot)$ is an indicator function. Figure~\ref{fig:model} (left) illustrates this with highlighted words.

\noindent\textbf{Entity Description Generation (EDG).}
EDG tasks the model with producing a text description for a given entity. For a text $X$ and associated entity set $E_{X}$, we construct a prefix text using the template ``Entities:'', followed by a list of entities and the template ``Text:''. The original text $X$ serves as the suffix. This forms the modified example $\hat{X}$ and corresponding label mask vector $\mathcal{M}_{\hat{X}}$, where $\mathcal{M}_{\hat{X}_i}=1$ if $\hat{x}_i$ is part of the suffix string.

\noindent\textbf{Knowledgeable Question Answering (KQA).}
KQA leverages relation triples from the KB to facilitate question answering. Given a text $X$ and entity set $E_X$, we select a pair of entities $e_h, e_t\in E_X$ linked by a 1-hop relation $r\in\mathcal{R}$ to form a triple $(e_h, r, e_t)\in\mathcal{T}$. Inspired by \citet{Wang2022Knowledge}, we create a question template for each triple, prompting the model to predict the tail entity $e_t$. Training examples $\hat{X}$ and label mask vectors are generated accordingly, with $\mathcal{M}_{\hat{X}_i}=1$ designating tokens belonging to the tail entity.

During pre-training, we randomly compile examples from the same task into a training batch $\mathcal{X}=\{\hat{X}\}$ until the maximum sequence length is reached.
The cross-entropy loss for prediction positions (where $\mathcal{M}_{\hat{X}}=1$) is computed as follows:
\begin{equation}
\mathcal{L}=\frac{1}{|\mathcal{X}|}\sum_{\hat{X}\in\mathcal{X}}\frac{1}{T_{\hat{X}}}\sum_{\hat{x}_i\in\hat{X}}\mathcal{M}_{\hat{X}_i}\log p(y_i|\hat{X}_{<i}),
\label{eqn:pretrain_loss}
\end{equation}
where $y_i$ is the ground truth token, $p(\cdot)$ is the predicted probability, and $T_{\hat{X}}=\sum_{\hat{x}_i\in\hat{X}}\mathcal{M}_{\hat{X}_i}$ is the number of tokens the model is required to predict.

\subsection{Knowledgeable Example Retrieval}
Despite having a powerful and knowledgeable LLM at our disposal, the efficacy of ICL is significantly influenced by the selection and ordering of labeled examples~\cite{Brown2020Language}. Previous studies~\cite{Liu2022What, Yao2022Fantastically, Rubin2022Learning} have demonstrated that LLMs can autonomously generate suitable text-based prompts, yet they largely overlook the importance of \emph{factual knowledge} from KBs.
To address this gap, we introduce a novel Knowledgeable Example Retrieval (KER) algorithm that utilizes knowledge to select the most relevant in-context examples. This process is illustrated in Figure~\ref{fig:model} (middle) and detailed in Algorithm~\ref{alg:ker} in Appendix~\ref{appendix:implementation_details}.
Concisely, given a training set $D_{trn}=\{(X_i^{trn}, y_i^{trn}, E_i^{trn})\}$ and a testing set $D_{tgt}=\{(X_j^{tgt}, E_j^{tgt})\}$, where $X_i^{trn}$ and $X_j^{tgt}$ are input texts, $y_i^{trn}$ are labels, and $E_i^{trn}$ and $E_j^{tgt}$ are the corresponding entity sets, KER's objective is to select a subset of training examples as demonstrations that exhibit high knowledge relevance to the testing set.

A straightforward approach is to retrieve examples containing entities that \emph{cover} a higher number of target examples. We use the Jaccard similarity to assess the similarity between two examples: 
\begin{equation}
d_{jac}(i, j)=\frac{|E_i^{trn}\cap E^{tgt}_j|}{|E_i^{trn}\cup E^{tgt}_j|}.
\end{equation}
However, since the Jaccard similarities for most example pairs are zero, we further employ pre-trained knowledge embeddings to retrieve training examples that are semantically \emph{similar} to the target set.
We compute the average representations $\mathbf{e}_i$ and $\mathbf{e}_j$ of all entities in $E^{trn}_i$ and $E^{tgt}_j$, respectively. The semantic difference is quantified using the Euclidean distance $d_{sem}(i, j)$ between $\mathbf{e}_i$ and $\mathbf{e}_j$.
The overall knowledge relevance between two examples is calculated as follows:
\begin{equation}
\begin{aligned}
&d(X_i^{trn}, X_j^{tgt}) =\alpha\frac{d_{jac}(i, j) + \gamma}{\max_{X_k^{trn}\in\mathcal{D}_{trn}}d_{jac}(i, k)+\gamma}\\
&+(1-\alpha)(1-\frac{d_{sem}(i, j)}{\max_{X_k^{trn}\in\mathcal{D}_{trn}}d_{sem}(i, k)}),
\label{eqn:relevance_score}
\end{aligned}
\end{equation}
where $\alpha\in[0,1]$ and $\gamma>0$ are tunable hyperparameters. The sampling weight for each training example $X_i^{trn}$ is given by:
\begin{equation}
\begin{aligned}
s'(X_i^{trn})=\frac{s(X_i^{trn})}{\sum_{X_j^{trn}\in\mathcal{D}_{trn}}s(X_j^{trn})},
\label{eqn:sample_weight}
\end{aligned}
\end{equation}
where $s(X_i^{trn})$ is computed as the average relevance score to the testing set:
\begin{equation}
\begin{aligned}
s(X_i^{trn})=\frac{1}{|\mathcal{D}_{tgt}|}\sum_{X_j^{tgt}\in\mathcal{D}_{tgt}}d(X_i^{trn}, X_j^{tgt}).
\label{eqn:average_score}
\end{aligned}
\end{equation}
An example with a higher weight signifies greater knowledge relevance across all target examples.
Ultimately, we sample $K$ training examples based on these weights to serve as in-context examples.


\subsection{Knowledgeable Prediction Calibration}
Following model pre-training and in-context example selection, we can proceed to generate predictions for the target example $X^{tgt}\in\mathcal{D}_{tgt}$ using the following equation:
\begin{equation}
\begin{aligned}
\hat{y} = \arg\max_{v\in\mathcal{V}} p(y=e|X, X^{tgt}),
\end{aligned}
\label{eqn:predict_proba}
\end{equation}
where $\mathcal{V}$ is a verbalizer that maps label words to their corresponding classes~\footnote{For classification tasks, $\mathcal{V}$ is the set of label words; for question answering tasks, $\mathcal{V}$ is the entire vocabulary.}. $\widetilde{\mathcal{D}}$ represents the set of in-context examples used for prediction. 
However, as discussed in Section~\ref{sec:preliminary}, the frequency of label words (in classification tasks) or entities (in question answering tasks) can bias the prediction probabilities. To mitigate this issue, we utilize the prior information of label words to refine the prediction for each target example.

Specifically, we select a subset of training data $\mathcal{S}$ from the KQA task and estimate the contextual prior probability for each candidate label word or entity $v\in\mathcal{V}$ at the output position:
\begin{equation}
\begin{aligned}
P(v)\approx\frac{1}{|\mathcal{S}|}\sum_{\hat{X}\in\mathcal{S}}p(y=v|\hat{X}),
\end{aligned}
\label{eqn:context_prior}
\end{equation}
where $\hat{X}$ denotes a training example, and $P(v)$ represents the estimated prior probability of candidate $v$.
Following this, we discard any label word or entity $v$ whose prior probability falls below a specific threshold~\cite{Hu2022Knowledgeable}.

Consequently, we enhance the final output by applying calibrated prediction:
\begin{equation}
\begin{aligned}
\hat{y} = \arg\max_{v\in\mathcal{V}} \frac{p(y=v|\widetilde{\mathcal{D}}, X^{tgt})}{P(v)}.
\end{aligned}
\label{eqn:calibrate_prediction}
\end{equation}

\noindent\textbf{Remarks}. While most related works~\cite{Hu2022Knowledgeable, Zhao2021Calibrate} concentrate on prediction calibration, our approach distinguishes itself by leveraging a priori knowledge from a large-scale corpus to debias outputs. This contrasts with methods that rely solely on in-domain data or utilize task-agnostic, content-free inputs (e.g., ``N/A'').



\section{Experiments}

\subsection{Implementation Settings and Baselines}
For the pre-training corpus, we use Wikipedia Dumps (2020/03/01)\footnote{\url{https://dumps.wikimedia.org/enwiki/}}, which consists of 25,933,196 sentences. Further, the KB we used is WikiData5M~\cite{Wang2021KEPLER}, which includes $3,085,345$ entities and $822$ relation types.
By default, we choose GPT-2 (large) with 0.8B parameters as the backbone.
For downstream tasks, we consider 8 text classification tasks and 4 question answering tasks. 
The details of corpora and downstream benchmarks are shown in Appendix~\ref{appendix:corpora_benchmark}. The implementation details of pre-training, prompting, and prediction can be found in Appendix~\ref{appendix:implementation_details}.


We consider the following baselines: 
1) \textbf{In-Context Learning (ICL)} is the vanilla version proposed by GPT-3. 
2) \textbf{Calibrate Before Use (CBU)}~\cite{Zhao2021Calibrate} is a typical method that aims to de-bias the prediction via content-free prompts. 
3) \textbf{KATE}~\cite{Liu2022What} uses the CLS embeddings of a RoBERTa-large model as sentence representations, and retrieves the nearest $K$ neighbors for each target example as the final in-context examples.
4) \textbf{MetaICL}~\cite{Min2022MetaICL} improves ICL by meta-learning the objective of ICL in cross-task settings.
5) \textbf{SelfSup.}~\cite{Chen2022Improving} improves ICL by multiple self-supervised learning tasks.
We also choose RoBERTa-large to perform fully \textbf{Fine-tuning} to demonstrate the ceiling performance of each task.

\begin{table*}[t]
\centering
\resizebox{\linewidth}{!}{
\begin{tabular}{lcccccccccc}
\toprule
\bf \multirow{2}*{\bf Baselines} & \bf  SST-2 & \bf  MRPC & \bf  MNLI & \bf  QNLI & \bf  RTE & \bf CB & \bf TREC & \bf AGNews & \bf \multirow{2}*{\bf Avg.} \\
& acc & f1 & acc & acc & acc & acc & acc & acc & \\
\midrule
\multicolumn{10}{l}{\textit{\textbf{Full Data}}}\\
Fine Tuning (RoBERTa-large) & 95.00 & 91.40 & 89.80 & 93.30 & 80.90 & 90.50 & 97.40 & 94.70 & 91.63 \\
\midrule
\multicolumn{10}{l}{\textit{\textbf{Few-shot Labeled Data (8-shot)}}}\\
ICL~\cite{Brown2020Language} & 
76.18\small{\textpm7.2} & 54.46\small{\textpm2.3} & 56.85\small{\textpm2.4} & 52.93\small{\textpm3.2} & 53.94\small{\textpm5.0} & 42.50\small{\textpm1.8} & 51.56\small{\textpm4.1} & 45.67\small{\textpm6.6} & 54.26 \\
CBU~\cite{Zhao2021Calibrate} & 
82.71\small{\textpm4.4} & 63.07\small{\textpm3.9} & 57.93\small{\textpm2.8} & 53.19\small{\textpm3.9} & 54.87\small{\textpm2.8} & 51.34\small{\textpm1.7} & 54.61\small{\textpm3.7} & 55.42\small{\textpm2.8} & 59.14 \\
KATE~\cite{Liu2022What} & 
81.33\small{\textpm3.8} & 58.04\small{\textpm3.9} & 59.40\small{\textpm2.4} & 53.57\small{\textpm3.5} & 53.17\small{\textpm2.7} & 45.48\small{\textpm2.1} & 54.69\small{\textpm2.8} & 50.28\small{\textpm3.4} & 57.00 \\
MetaICL$^{\dag}$~\cite{Min2022MetaICL} & 
87.40\small{\textpm5.0} & 62.91\small{\textpm2.0} & 60.22\small{\textpm3.4} & 55.18\small{\textpm1.9} & 57.06\small{\textpm2.8} & 49.20\small{\textpm2.5} & 56.09\small{\textpm1.8} & 55.80\small{\textpm2.4} & 60.48 \\
SelfSup.$^{\dag}$~\cite{Chen2022Improving} & 
87.94\small{\textpm3.0} & 62.33\small{\textpm2.0} & 62.00\small{\textpm2.2} & 54.77\small{\textpm1.8} & 57.27\small{\textpm2.6} & 45.80\small{\textpm2.5} & 55.59\small{\textpm2.5} & 57.44\small{\textpm3.2} & 60.39 \\
\hdashline
{\model}$^{\dag}$ & \bf
91.21\small{\textpm2.9} & \bf 69.96\small{\textpm0.7} & \bf 69.59\small{\textpm1.0} & \bf 60.66\small{\textpm1.2} & \bf 63.74\small{\textpm4.2} & \bf 56.07\small{\textpm3.8} & \bf 63.52\small{\textpm5.5} & \bf 68.89\small{\textpm5.7} & \bf 67.96 \\
\qquad only w. KPT$^{\dag}$ &
90.04\small{\textpm3.5} & 66.65\small{\textpm1.9} & 67.39\small{\textpm2.6} & 58.97\small{\textpm3.0} & 58.26\small{\textpm3.3} & 55.43\small{\textpm2.0} & 60.16\small{\textpm2.2} & 59.74\small{\textpm4.4} & 64.58 \\
\qquad only w. KER &
84.05\small{\textpm2.7} & 59.26\small{\textpm2.5} & 59.93\small{\textpm1.0} & 57.23\small{\textpm1.2} & 53.79\small{\textpm4.0} & 51.36\small{\textpm3.8} & 55.52\small{\textpm5.1} & 52.70\small{\textpm3.3} & 59.23 \\
\qquad only w. KPC &
85.52\small{\textpm3.9} & 64.77\small{\textpm0.7} & 63.13\small{\textpm1.2} & 57.69\small{\textpm2.4} &
55.94\small{\textpm1.2} & 54.07\small{\textpm2.8} & 56.92\small{\textpm2.7} & 57.24\small{\textpm5.5} & 61.91 \\

\bottomrule
\end{tabular}
}
\caption{The 8-shot performance (\%) on GPT-2 (large) of different learning settings with standard deviations over text classification benchmarks. Compared with other baselines, our framework achieves consistent improvement. $^{\dag}$ denotes the method involves parameters update for ICL. ``only w.'' means we only use one technique in~{\model}.}
\label{tab:main-result-cls}
\end{table*}

\subsection{Main Results}
Table~\ref{tab:main-result-cls} and Table~\ref{tab:main-result-qa} respectively report the
results over text classification and question answering tasks in the 8-shot setting. We thus make the following observations:
1) Our proposed framework outperforms strong baselines and achieves substantial improvements over all benchmarks.
Specifically, compared with ICL, the average result over the text classification task is improved by 13.70\%, which is larger than that of other baselines.
The average gain over question answering tasks is also more than 7\%, although there is still room for improvement on unseen target domains, likely because they require more challenging generalization and commonsense abilities.
2) Compared with ICL, KER and KCP make significant contributions to the performance. Particularly, KER and KCP also respectively outperform strong baselines KATE and CBU, indicating the indispensable merit of factual knowledge at the inference stage. 
3) The performance of KPT exceeds that of meta-learning (MetaICL) and self-supervised learning (SelfSup.) approaches by around 4\%, which also focus on continual pre-training. 
This demonstrates that explicitly injecting knowledge into LLMs is more effective for ICL, which is imperative and plays a dominant role. 
4) Our method attains more impressive performance when combining all of these knowledgeable techniques, highlighting the necessity of factual knowledge in ICL. We provide a detailed analysis in Section~\ref{ablation_study}.
5) We also evaluate other scales for GPT-2 and OPT in 8-shot settings. Results in Appendix~\ref{appdix:more_results} show that the improvements are consistent across different LLMs.

\begin{table*}[t]
\centering
\begin{small}
\begin{tabular}{lcccccc}
\toprule
\bf \multirow{2}*{\bf Baselines} & \bf  ComQA & \bf  Quartz & \bf  SQuAD & \bf  Quoref & \bf \multirow{2}*{\bf Avg.} \\
& acc & acc & em & em & \\
\midrule
\multicolumn{6}{l}{\textit{\textbf{Full Data}}}\\
Fine Tuning (RoBERTa-large) & 72.10 & 76.90 & 86.50 & 78.70 & 78.55 \\
\midrule
\multicolumn{6}{l}{\textit{\textbf{Few Labeled Data (8-shot)}}}\\
ICL~\cite{Brown2020Language} & 
27.93\small{\textpm4.8} & 54.49\small{\textpm3.5} & 46.93\small{\textpm3.0} & 40.31\small{\textpm2.7} & 42.42 \\
CBU~\cite{Zhao2021Calibrate} & 
29.88\small{\textpm3.9} & 55.40\small{\textpm1.8} & 49.32\small{\textpm4.0} & 44.05\small{\textpm4.0} & 44.66 \\
KATE~\cite{Liu2022What} & 
29.02\small{\textpm4.0} & 55.10\small{\textpm3.9} & 47.25\small{\textpm3.4} & 42.77\small{\textpm3.8} & 43.54 \\
MetaICL$^{\dag}$~\cite{Min2022MetaICL} & 
31.16\small{\textpm3.2} & 55.64\small{\textpm2.9} & 50.46\small{\textpm2.6} & 46.72\small{\textpm2.7} & 46.00 \\
SelfSup.$^{\dag}$~\cite{Chen2022Improving} & 
31.32\small{\textpm3.0} & 54.88\small{\textpm3.0} & 49.97\small{\textpm2.7} & 47.50\small{\textpm3.5} & 45.92 \\
\hdashline
{\model}$^{\dag}$ & 
\bf 36.17\small{\textpm1.8} & \bf 58.11\small{\textpm2.4} & \bf 54.23\small{\textpm2.6} & \bf 50.46\small{\textpm3.3} & \bf 49.74 \\
\qquad only w. KPT$^{\dag}$ & 
34.21\small{\textpm4.3} & 57.32\small{\textpm2.2} & 52.79\small{\textpm3.0} & 49.93\small{\textpm1.9} & 48.56 \\
\qquad only w. KER & 
29.56\small{\textpm2.3} & 55.82\small{\textpm1.2} & 48.11\small{\textpm2.4} & 43.58\small{\textpm2.1} & 44.27 \\
\qquad only w. KCP & 
33.60\small{\textpm3.7} & 57.77\small{\textpm2.4} & 51.63\small{\textpm2.9} & 46.09\small{\textpm3.1} & 47.27 \\

\bottomrule
\end{tabular}
\end{small}
\caption{The 8-shot performance (\%) on GPT-2 (large) of different learning settings with standard deviations over question answering benchmarks.}
\label{tab:main-result-qa}
\end{table*}

\begin{table*}[t]
\centering
\resizebox{\linewidth}{!}{
\begin{tabular}{lccccccccccc}
\toprule
\bf \multirow{2}*{\bf Baselines} & \bf  SST-2 & \bf MRPC & \bf MNLI & \bf RTE & \bf  AGNews & \bf  TREC & \bf  ComQA & \bf Quartz & \bf SQuAD & \bf Quoref \\
& acc & f1 & acc & acc & acc & acc & acc & acc & em & em \\
\midrule
ICL & 
76.18\small{\textpm7.2} & 
54.46\small{\textpm2.3} & 
56.85\small{\textpm2.4} & 
53.94\small{\textpm5.0} & 
45.67\small{\textpm6.6} & 
51.56\small{\textpm4.1} & 
27.93\small{\textpm4.8} & 
54.49\small{\textpm3.5} & 46.93\small{\textpm3.0} & 40.31\small{\textpm2.7} \\
\hdashline
KPT+KER & 
\underline{91.04\small{\textpm3.3}} & 
67.93\small{\textpm3.0} & 
68.47\small{\textpm2.9} & 
61.30\small{\textpm3.3} & 
62.18\small{\textpm3.9} & 
61.52\small{\textpm3.1} & 
35.17\small{\textpm4.0} & 
57.64\small{\textpm2.6} & 
52.23\small{\textpm3.4} &
\underline{50.20\small{\textpm3.1}} \\
KPT+KCP & 
90.65\small{\textpm3.7} &
\underline{68.44\small{\textpm2.5}} & 
\underline{68.89\small{\textpm3.4}} & 
\underline{62.38\small{\textpm2.3}} & 
\underline{63.88\small{\textpm3.5}} & 
\underline{62.12\small{\textpm2.9}} & 
\bf 36.38\small{\textpm2.2} & 
\underline{58.03\small{\textpm2.0}} & 
\underline{54.17\small{\textpm1.8}} &
50.18\small{\textpm2.2} \\
KER+KCP & 
86.45\small{\textpm3.0} & 
64.07\small{\textpm2.4} & 
66.60\small{\textpm2.9} & 
57.39\small{\textpm3.2} & 
58.95\small{\textpm3.6} & 
58.60\small{\textpm3.5} & 
34.26\small{\textpm2.2} & 
57.88\small{\textpm3.1} & 
52.20\small{\textpm2.3} & 47.92\small{\textpm2.7} \\
\hdashline
All ({\model}) & 
\bf 91.21\small{\textpm2.9} & 
\bf 69.96\small{\textpm0.7} & 
\bf 69.59\small{\textpm1.0} & 
\bf 63.74\small{\textpm4.2} & 
\bf 68.89\small{\textpm5.7} & 
\bf 63.52\small{\textpm5.5} & 
\underline{36.17\small{\textpm1.8}} & 
\bf 58.11\small{\textpm2.4} & 
\bf 54.23\small{\textpm2.6} & 
\bf 50.46\small{\textpm3.1} \\
\bottomrule
\end{tabular}
}
\caption{The 8-shot performance (\%) of different combinations of the knowledgeable modules.}
\label{tab:ablation}
\end{table*}

\begin{table}[t]
\centering
\resizebox{\linewidth}{!}{
\begin{tabular}{lcccccc}
\toprule
\bf \multirow{2}*{\bf Methods} & \bf  SST-2 & \bf  AGNews & \bf  TREC & \bf  ComQA & \bf SQuAD \\
& acc & acc & acc & acc & em \\
\midrule
None (ICL) & 
76.18\small{\textpm7.2} & 
45.67\small{\textpm6.6} & 
51.56\small{\textpm4.1} & 
27.93\small{\textpm4.8} & 46.93\small{\textpm3.0} \\
GPT-2 & 
81.35\small{\textpm}3.0 & 
48.72\small{\textpm}2.7 & 
52.36\small{\textpm}3.3 & 
28.61\small{\textpm}3.8 &
47.14\small{\textpm}3.1 \\
\hdashline
KPT & 
\bf 90.04\small{\textpm3.5} & 
\bf 59.74\small{\textpm4.4} & 
\bf 60.16\small{\textpm2.0} & 
\bf 34.21\small{\textpm4.3} & 
\bf 52.79\small{\textpm3.0} \\
\quad w/o. MEP & 
84.40\small{\textpm4.0} & 
51.29\small{\textpm3.9} & 
54.72\small{\textpm3.1} & 
\underline{33.01\small{\textpm7.7}} &
\underline{52.23\small{\textpm2.8}} \\
\quad w/o. EDG & 
\underline{87.19\small{\textpm2.9}} & 
\underline{56.40\small{\textpm4.3}} & 
\underline{55.91\small{\textpm3.1}} & 
31.95\small{\textpm5.9} &
50.80\small{\textpm3.9} \\
\quad w/o. KQA & 
85.30\small{\textpm3.3} & 
53.03\small{\textpm3.6} & 
53.46\small{\textpm2.4} & 
30.08\small{\textpm5.8} &
49.71\small{\textpm4.6} \\
\bottomrule
\end{tabular}
}
\caption{The 8-shot performance (\%) of each self-supervised task. GPT-2 denotes the vanilla objective.}
\label{tab:result_kpt}
\end{table}

\subsection{Ablation Study}
\label{ablation_study}

We further investigate how these proposed knowledgeable techniques contribute to the final performance with different combinations. As shown in Table~\ref{tab:ablation}, the results demonstrate that any combination greatly promotes the overall performance of vanilla ICL. An interesting observation is that KPT is particularly important for performance improvement, achieving higher scores than KER and KCP. This indicates that the most effective way to unleash the power of LLMs is to inject knowledge into the model parameters. Nonetheless, the combination of KER and KCP also improves ICL by about 8\% for each task, respectively. This suggests that KER and KCP are critical to ICL because ultra-large LLMs cannot be continuously pre-trained or tuned in real-world scenarios to save computational resources.
Furthermore, results from Table~\ref{tab:main-result-cls} to Table~\ref{tab:ablation} show that our method has significantly improved classification tasks. We believe that the benefits of injecting knowledge are more pronounced for simple language understanding tasks than for question answering.


\begin{figure}[t]
\centering
\includegraphics[width = 0.65\linewidth]{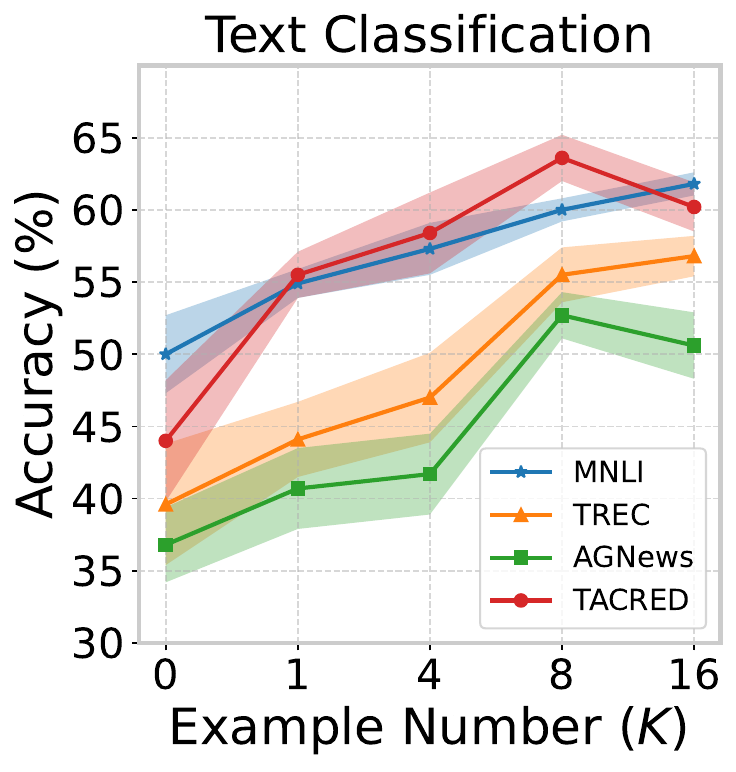}
\includegraphics[width = 0.65\linewidth]{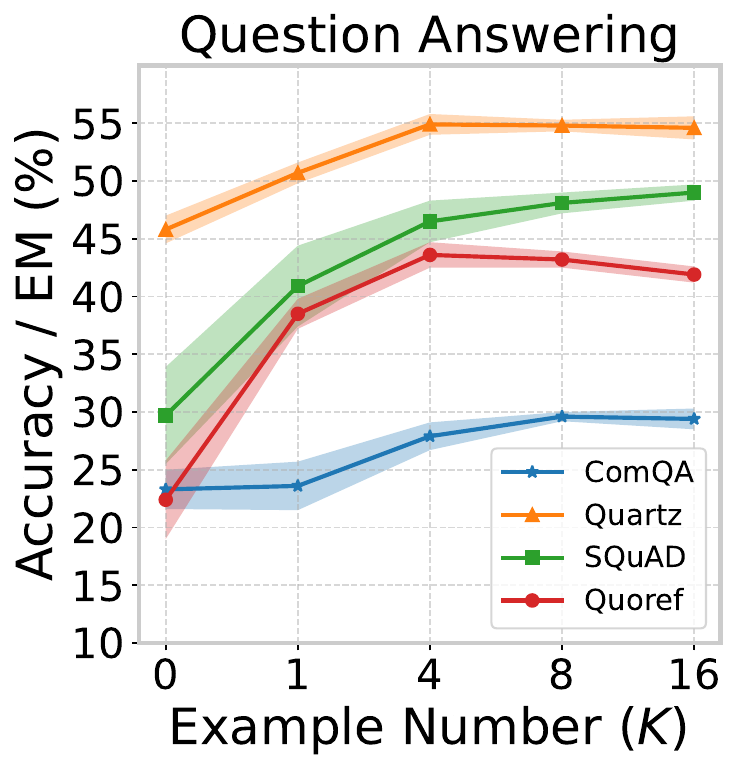}
\caption{GPT-2 (large) sample effectiveness (\%) of~{\model} (only w. KER) with different values of $K$.
}
\label{fig:result_ker1}
\end{figure}

\subsection{Further Analysis}
\label{further_analysis}

\paragraph{Effectiveness of KPT.}
To investigate what makes a high performance for KPT, we test the effectiveness of each knowledgeable self-supervised task. 
For a fair comparison, we also choose two baselines:
1) \textbf{None} is that we do not use any self-supervised task, which is the same as vanilla ICL proposed in~\cite{Brown2020Language},
2) \textbf{GPT-2} represents conventional autoregressive language modeling (ALM) pre-training tasks.
As shown in Table~\ref{tab:result_kpt}, 
KPT can make substantial improvements for ICL.
Particularly, all the self-supervised learning tasks in KPT are complementary for pre-training and outperform the baseline with or without the conventional objective of GPT-2.
In addition, the MEP and KQA tasks are most critical for classification and question answering, respectively, which demonstrates that different pre-training objectives possess different advantages in downstream tasks.

\begin{figure}[t]
\centering
\begin{minipage}[t]{0.49\linewidth}
    \includegraphics[width = 1\linewidth]{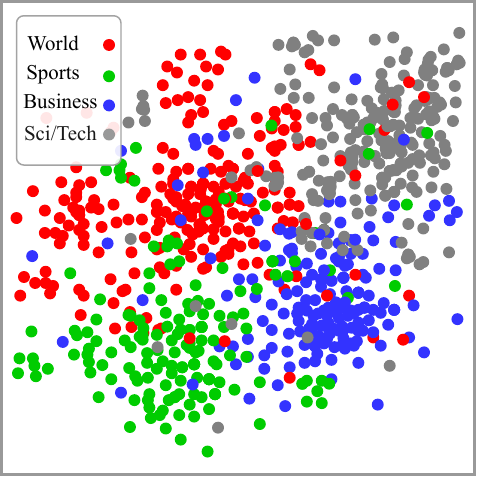}
\end{minipage}
\begin{minipage}[t]{0.49\linewidth}
    \includegraphics[width = 1\linewidth]{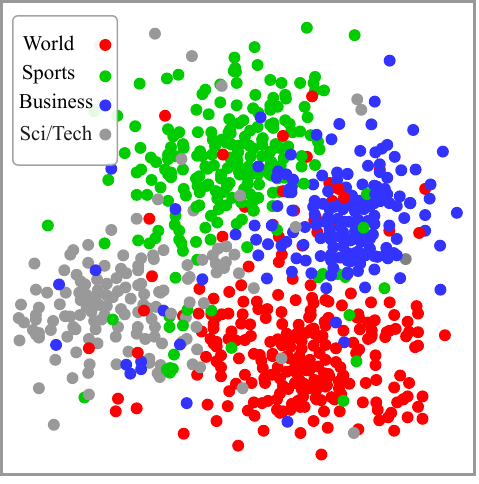}
\end{minipage}
\caption{Visualizations of each AGNews's training example. KATE (left) uses CLS embeddings of RoBERTa. Ours (right) utilizes averaged knowledge embeddings.}
\label{fig:visualization}
\end{figure}

\paragraph{Sample Effectiveness.}
To investigate the influence of the number of in-context examples $K$, we choose multiple classification and question answering tasks and vary $K$ from 0, 1, 4, 8 to 16.
From Figure~\ref{fig:result_ker1}, we find that increasing $K$ generally helps across both classification and question answering tasks, demonstrating that more in-context examples may bring more knowledge to better guide the LLM to make predictions.
When $K>8$, the performance of the most tasks will decrease, because the maximum length limit causes information loss. 
The suitable value $K$ is set around 8.

\paragraph{Visualization of Selected Examples in KER.}
In addition, for explicitly seeing the performance in semantic space, we obtain the t-SNE~\cite{van2008visualizing} visualization of each training example over AGNews via averaged representations of all corresponding entities. 
We choose KATE as our strong baseline, which is also focused on the example selection. Here, we do not fine-tune RoBERTa on the training set. Figure~\ref{fig:visualization} demonstrates that our method can build better semantic representations toward factual knowledge.

\paragraph{Permutations of In-Context Examples.}
We also compare different permutations of these selected examples according to the sample weight computed in Eq.~\ref{eqn:sample_weight}.
In Table~\ref{tab:result_ker_2}, Random means to randomly choose an order. Ascending and Descending respectively denote that the example order is ascending or descending by weight.
From the results, we find no tangible relationship between the sampling weight and order.

\begin{figure}[t]
\centering
\begin{minipage}[t]{0.49\linewidth}
    \includegraphics[width = 1\linewidth]{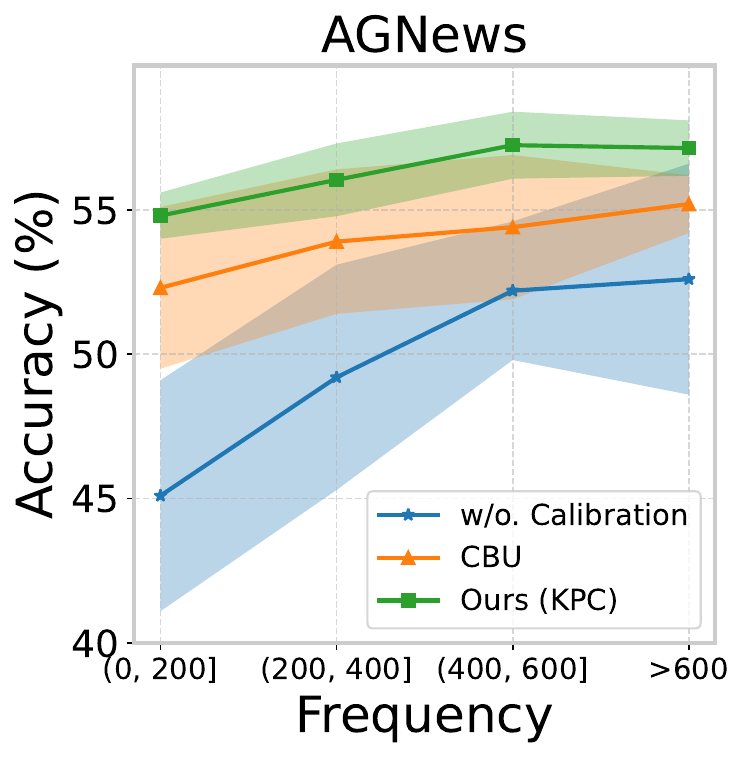}
\end{minipage}
\begin{minipage}[t]{0.49\linewidth}
    \includegraphics[width = 1\linewidth]{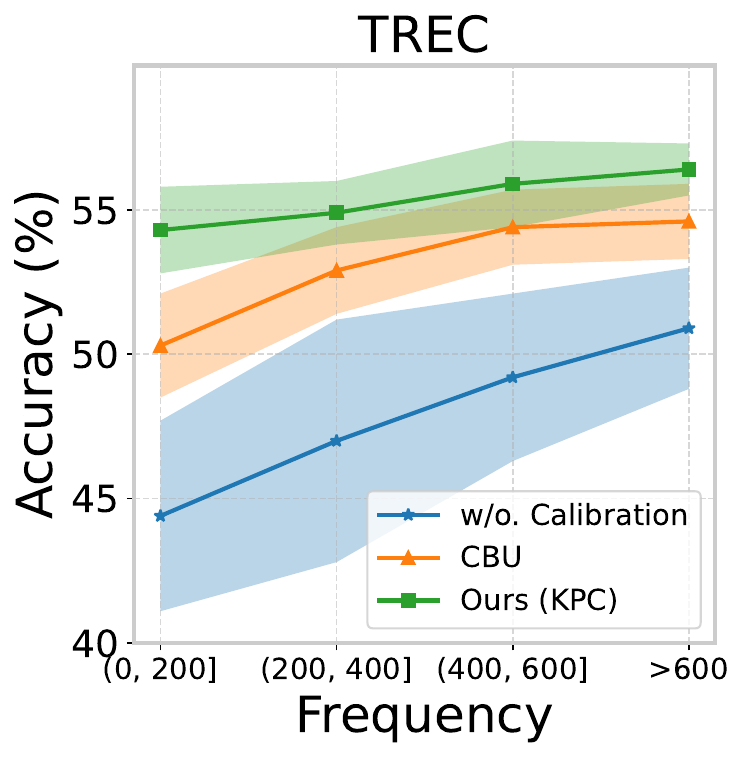}
\end{minipage}
\caption{GPT-2 (large) 4-shot performance of calibration over difference word frequencies.}
\label{fig:result_kcp}
\end{figure}

\paragraph{Effectiveness of KPC.}
We finally conduct analysis on prediction calibration. We choose AGNews and TREC tasks and follow the same settings in the preliminary experiments (we randomly choose two label words from different frequency regions).
Results in Figure~\ref{fig:result_kcp} demonstrate that calibrating the prediction consistently achieves improvements to the vanilla approach. 
In addition, we find that the prediction results highly depend on the label frequency, which is similar to Figure~\ref{fig:preexp2}.
However, our KPC still outperforms the strong baseline Calibrate Before Use (CBU) with arbitrary label frequency, which only transforms the input into content-free prompts.
It underscores that the prior information of each label word in KB is non-negligible.
In other words, calibration by the prior information can alleviate the impact of label frequency. 

\begin{table}[t]
\centering
\begin{small}
\begin{tabular}{lccc}
\toprule
Baselines & SST-2 & MRPC & MNLI \\
\midrule
Random & 
79.42\small{\textpm2.7} & 
\bf 59.26\small{\textpm2.5} & 
\bf 59.93\small{\textpm1.0} \\
Ascending & 
78.29\small{\textpm2.2} & 
58.05\small{\textpm2.6} & 
59.31\small{\textpm1.5} \\
Descending & 
\bf 79.61\small{\textpm3.0} & 
58.16\small{\textpm3.0} & 
59.58\small{\textpm1.3} \\
\bottomrule
\end{tabular}
\end{small}
\caption{The 8-shot averaged results (\%) of {\model} (only w. KER) for different permutations.}
\label{tab:result_ker_2}
\end{table}

\section{Related Work}

\subsection{Pre-trained/Large Language Models}
Pre-trained Language Models (PLMs) aim to learn representations from texts and have made significant progress in NLP. PLMs can be divided into three main categories: encoder-only~\cite{Devlin2019BERT, Liu2019RoBERTa, He2021Deberta, Yang2019XLNet, Lan2020ALBERT,DBLP:conf/emnlp/ZhangDWWWLHLH22}, decoder-only~\cite{radford2018improving, Brown2020Language, Zhang2022OPT}, and encoder-decoder~\cite{Lewis2020BART, Raffel2020Exploring}.
To incorporate factual knowledge into PLMs, a branch of knowledge-enhanced PLMs has been proposed~\cite{Zhang2019ERNIE, Sun2020CoLAKE, Wang2021KEPLER, Wang2021KAdapter, Wang2022Knowledge, Pan2022Knowledge,DBLP:conf/aaai/Zhang0HQTH022}, enabling PLMs to capture rich semantic knowledge from KBs. 
Since the introduction of ChatGPT, a variety of decoder-only LLMs have been released. Popular open-source LLMs include LLaMA~\cite{DBLP:journals/corr/abs-2307-09288}, OPT~\cite{Zhang2022OPT}, 
Galactica~\cite{DBLP:journals/corr/abs-2211-09085}, Pythia~\cite{DBLP:conf/icml/BidermanSABOHKP23}, among others.
Our work concentrates on decoder-only LLMs and aims to infuse them with factual knowledge to enhance their ICL performance.

\subsection{Prompt Learning}
Prompt-based learning aims to add natural language prompts to guide PLMs to solve downstream tasks.
A series of works focus on tunable discrete prompt tuning~\cite{Gao2021Making, Raffel2020Exploring} and continuous prompt tuning~\cite{Xiao2021GPT, Gu2021PPT, DBLP:conf/wsdm/Xu0QLXHH23}. 
For LLMs, GPT-3~\cite{Brown2020Language} enables In-Context Learning (ICL) with a text-based prompt in zero-shot scenarios, bypassing parameter updates~\cite{Dong2023A}.
To explore the factors affecting ICL, previous works have focused on input-output mapping~\cite{Min2022Rethinking, Kim2022Ground}, meta-learning~\cite{Chen2022Meta, Min2022MetaICL}, prompt engineering~\cite{Liu2022What, Liu2021Pre}, and prediction calibration~\cite{Zhao2021Calibrate, Hu2022Knowledgeable}, among others.
Recently, the Chain-of-Thought (CoT) approach has been presented to leverage reasoning and interpretable information to guide LLMs in generating reliable responses~\cite{Si2022Prompting, Zhang2022Automatic, Wei2022Chain,DBLP:conf/emnlp/Yan0ZHHZ23}.
Different from these approaches, we exploit \emph{factual knowledge} to further improve ICL in pre-training, prompting, and prediction phases.

\section{Conclusion}
In this paper, we investigate and harness \emph{factual knowledge} in ICL, including inherent knowledge embedded in LLMs, pertinent knowledge derived from selected training examples, and knowledge biases affecting predictions.
We introduce a novel Knowledgeable In-Context Tuning (\textbf{KICT}) framework to further enhance ICL performance by comprehensively exploiting factual knowledge throughout the processes of pre-training, prompting, and prediction. Experiments demonstrate that each introduced technique significantly improves upon strong baselines across classification and question-answering tasks.
Future work will focuses on 1) exploring the reasoning capabilities and interpretability of knowledge within ICL, and 2) extending our approach to encoder-decoder models.

\section*{Acknowledgements}


This work has been supported by the National Natural Science Foundation of China under Grant No. U1911203, 
Alibaba Group through the Alibaba Innovation Research (AIR) Program, 
the National Natural Science Foundation of China under Grant No. 61877018,
the Research Project of Shanghai Science and Technology Commission (20dz2260300) and the Fundamental Research Funds for the Central Universities.

\section*{Limitations}
This work presents several limitations: 1) It concentrates on decoder-only LLMs, as traditional in-context learning primarily targets decoder-only generation models such as GPT-2, GPT-3, OPT, etc. Nevertheless, we envision potential extensions to encoder-decoder architectures used in tasks such as translation and conditional generation.
2) Due to computational resource constraints, we do not experiment with ultra-large LLMs exceeding 10 billion parameters.
3) Our investigation centers on factual knowledge in three specific areas: pre-training, prompting, and prediction. We acknowledge that knowledge may influence additional aspects such as reasoning and interpretability, and we intend to explore these in future research.

\section*{Ethical Considerations}
The contributions of this work are methodological, focusing on a Knowledgeable In-Context Tuning (\textbf{KICT}) framework to augment the capabilities of LLMs with factual knowledge. Nonetheless, transformer-based models may perpetuate negative biases, including gender and social biases. As such, these issues are inherent to our work as well. We advise caution and recommend addressing potential risks when \textbf{KICT} models are deployed in real-world applications.





\appendix

\section{Details of Preliminary Experiments}
\label{appendix:preliminary}

\subsection{Details of Destruction Settings}
For our preliminary experiments, we selected 8 classification tasks and 4 question-answering tasks. The specifics of these datasets are detailed in Appendix~\ref{appendix:corpora_benchmark}.
To explore the influence of \emph{factual knowledge}, we posit that entities (and their associated labels in text classification tasks) embody factual knowledge~\cite{Wang2021KEPLER, Wang2022Knowledge, Wang2021KAdapter, Sun2019ERNIE, Zhang2019ERNIE}.
We identify all entities using the open-source TagMe entity linking tool\footnote{\url{https://sobigdata.d4science.org/group/tagme}}~\cite{Ferragina2010TAGME}.
In the case of classification tasks, labels are treated as special types of entities.
We follow the methodologies of \citet{Min2022Rethinking} and \citet{Kim2022Ground} to create various destruction settings that either remove or replace entities (and labels), thereby demonstrating the impact of factual knowledge.
Additionally, for each task, we randomly select $K=8$ examples as in-context examples and concatenate them with each test example to form an input sequence, capped at a maximum sequence length of 256 tokens.
With 5 different random seeds (i.e., 12, 24, 42, 90, and 100), each dataset yields 5 unique test results for a given LLM. Consequently, for each LLM, we collate $8 \times 5 = 40$ results for classification and $4 \times 5 = 20$ results for question-answering tasks.
The aggregated results are presented in Figure~\ref{fig:preexp1}, underscoring factual knowledge as a pivotal component in the performance of ICL.

\subsection{Details of Frequency Settings}
\begin{figure}
\centering
\includegraphics[width=\linewidth]{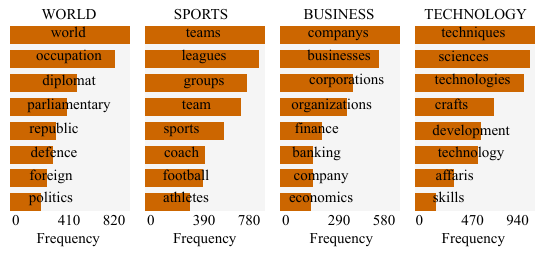}
\caption{Label word frequency statistics for the AGNews dataset.}
\label{fig:frequency_agnews}
\end{figure}

In our preliminary assessment of label word frequency's impact, we focused on two well-established tasks: AGNews and TREC.
Selecting $K=4$ examples from the training corpus to construct the in-context prompt, we then used the remaining training examples as targets to generate predictions. Development or test sets were not utilized due to their insufficient scale for demonstrating frequency effects clearly.
During prediction, we recorded the top-4 words with the highest prediction probabilities, facilitating the computation of frequency statistics for each label word.
Figure~\ref{fig:frequency_agnews} depicts the top-8 label word frequency statistics for each AGNews category.
To examine frequency influences, we randomly selected two label words per frequency range (e.g., $(0, 200]$, $(200, 400]$, $(400, 600]$, and $>600$) for predictions. For instance, in AGNews, labels like ``teams'' and ``groups'' could be chosen from the $>600$ frequency region to represent the ``sports'' category.
Accordingly, we generated $2^4=16$ and $2^6=64$ permutations for AGNews and TREC, respectively.
We report the average results using GPT-2 (urge) with 1.5B parameters and present the findings in box plot format in Figure~\ref{fig:preexp2}.

\subsection{Analysis of Knowledge Relevance in In-Context Examples}
\begin{figure}[t]
\centering
\includegraphics[width = 0.7\linewidth]{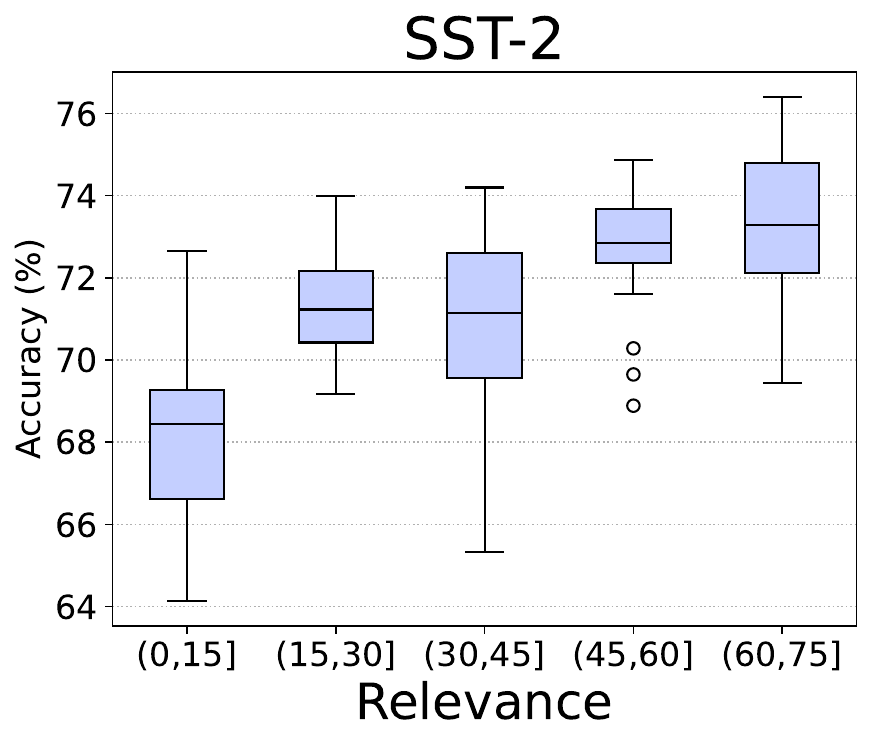}
\includegraphics[width = 0.7\linewidth]{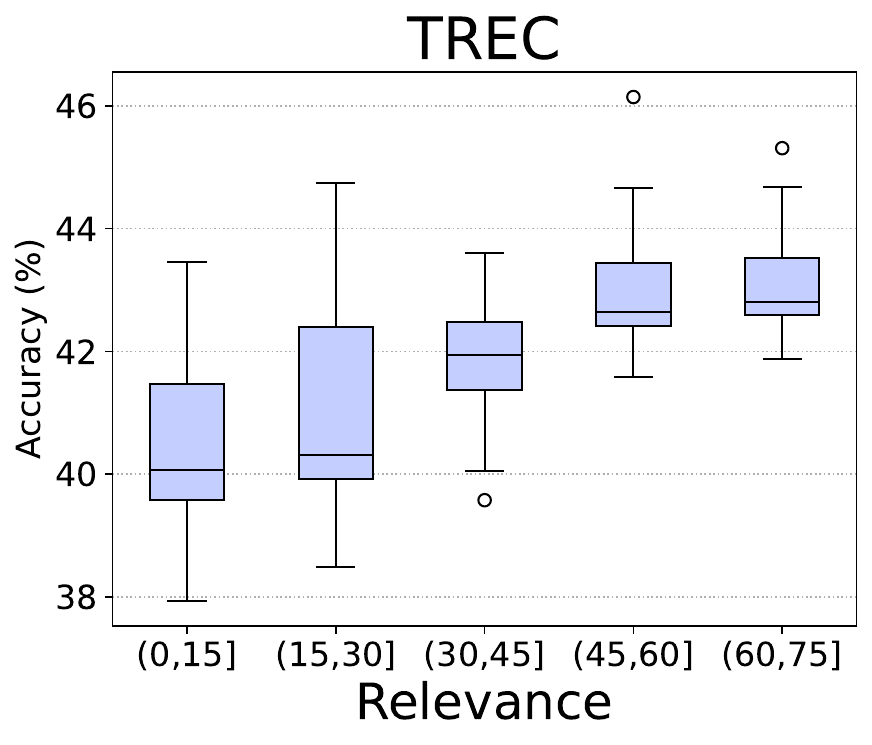}
\caption{The 4-shot performance (\%) with different knowledge relevance over SST-2 and TREC.}
\label{fig:preexp2_relevance}
\end{figure}

Our preliminary experiments indicated that factual knowledge in selected in-context examples is crucial for ICL. To substantiate this, we conducted further analyses on two datasets, SST-2 and TREC.
Employing our KER technique, we calculated a knowledge relevance score for each training example.
For each defined score interval (i.e., $(0, 15]$, $(15, 30]$, $(30, 45]$, $(45, 60]$, $(60, 75]$), we sampled $K=4$ examples to compose the in-context prompt.
We then assessed the average performance across all $4! = 24$ permutations for each interval and visualized the results in Figure~\ref{fig:preexp2_relevance}.
The findings corroborated the significance of selecting examples with high knowledge relevance for enhancing ICL performance.

\section{Details of the Corpus and Downstream Benchmarks}
\label{appendix:corpora_benchmark}

\subsection{Corpora and Knowledge Base}

We propose knowledgeable pre-training (KPT), which is similar to the current flourishing research of \emph{knowledge-enhanced pre-trained language models} (KEPLMs)~\cite{Liu2020KBERT, Sun2019ERNIE, Sun2020ERNIE2, Wang2022Knowledge}.
Different from them, we focus on auto-regressive PLMs, such as GPT-2.
We collect training corpora from Wikipedia (2020/03/01)\footnote{\url{https://dumps.wikimedia.org/enwiki/}.}, and use WikiExtractor\footnote{\url{https://github.com/attardi/wikiextractor}.} to process the pre-training data.
The knowledge base (KB) $\mathcal{G}$ we choose is WikiData5M~\cite{Wang2021KEPLER}, which is an urge-large structural data source based on Wikipedia.
The entity linking toolkit we used is TagMe.
In total, we have 3,085,345 entities and 822 relation types in $\mathcal{G}$, and 25,933,196 training sentences.

As mentioned above, KPT consists of three self-training tasks, i.e., \emph{masked entity prediction}, \emph{entity description generation}, and \emph{knowledgeable question answering}. 
For each task, we randomly select multiple sentences to form a training instance until reaching the maximum sequence length (i.e., 2048). 
Finally, we have sampled 100k training instances for each task.
In average, we have 8 examples for each instance.

\begin{table*}[t]
\begin{center}
\centering
\resizebox{\linewidth}{!}{%
\begin{tabular}{clcrrrcl}
\toprule
\bf Category & \bf Dataset & \bf \#Class & \bf \#Train & \bf \#Test & \bf Type & \bf Labels (classification tasks) \\
\bottomrule
 & SST-2 & 2 & 6,920 & 872 & sentiment & positive, negative \\
& MRPC & 2  & 3,668 & 408 & paraphrase & equivalent, not\_equivalent \\
& MNLI & 3 & 392,702 & 9,815 & NLI & entailment, neutral, contradiction\\ Text & QNLI & 2  & 104,743 & 5,463 & NLI & entailment, not\_entailment \\
Classification & RTE & 2 & 2,490 & 277 & NLI &  entailment, not\_entailment \\
& CB & 3 & 250 & 57 & NLI & entailment, neutral, contradiction \\
& TREC & 6 & 5,452 & 500 & question cls. & abbr., entity, description, human, loc., num.\\
& AGNews & 4 & 120,000 & 7,600 & topic cls. & world, sports, business, technology \\
\midrule
& ComQA & - & 9,741 & 1,221 & multi-choice & - \\
Question & Quartz & - & 2,696 & 384 & multi-choice & - \\
Answering & SQuAD & - & 87,599 & 10,570 & extractive QA & - \\
& Quoref & - & 19,399 & 2,418 & extractive QA & - \\
\bottomrule
\end{tabular}
}
\end{center}
\caption{The statistics of multiple text classification and question answering datasets. Since the original test data is unavailable, we use the development sets as our test sets.
}
\label{tab:datasets}
\end{table*}

\subsection{Downstream Task Datasets}
To evaluate the effectiveness of our framework, we choose 8 text classification tasks and 4 question answering tasks.
For the text classification, we directly choose 8 tasks from~\cite{Gao2021Making, Zhao2021Calibrate}. 
All the classification tasks involve sentiment analysis, natural language inference (NLI), question classification, and topic classification.
For the question answering tasks, we choose four widely used tasks, including CommonsenseQA (ComQA)~\cite{Talmor2019Alon}, Quartz~\cite{Tafjord2019QuaRTz}, SQuAD~\cite{Pranav2018Know} and Quoref~\cite{Dasigi2019Quoref}, where ComQA and Quartz are multi-choice QA, SQuAD and Quoref are extractive QA.
The statistics of each dataset are shown in Table~\ref{tab:datasets}.

\section{Implementation Details}
\label{appendix:implementation_details}
\subsection{Pre-training Details}

In the pre-training stage, we choose different scales of GPT-2 (0.1B, 0.3B, 0.8B, 1.5B)~\cite{Brown2020Language} and OPT~\cite{Zhang2022OPT} (2.7B, 6.7B) from HuggingFace\footnote{\url{https://huggingface.co/transformers/index.html}.} as the underlying LLMs. 
We do not use larger GPT-3 models because of the computation resource limitations.
Because all three kinds of pre-training tasks share the same format, we can directly mix up all the pre-training examples to form a cross-task pre-training paradigm. 
We find that it is suitable for the LLM to learn cross-task knowledge.
We train our model by AdamW algorithm with $\beta_1=0.9, \beta_2=0.98$. The learning rate is set as 1e-5 with a warm-up rate 0.1. 
We also leverage dropout and regularization strategies to avoid over-fitting. 
The models are trained on 8 NVIDIA A100-80G GPUs.

\subsection{Prompting Details}
We describe the implementation details with knowledgeable example retrieval (KER). Given a training dataset and a testing set, we aim to choose $K$ examples from the training set which have a high knowledge relevant to all testing examples. 
To reach this goal, we utilize both Jaccard similarity and Euclidean distance in terms of pre-trained knowledge embeddings.
For pre-trained knowledge embeddings, we choose the ConVE~\cite{Tim18Convolutional} algorithm to pre-train over wikidata5m and obtain the embeddings of entities and relations. We set its dimension as 768, the negative sampling size as 64, the batch size as 128 and the learning rate as 0.001. Finally, we only store the embeddings of all the entities.
The KER algorithm for the prompting is shown in Algorithm~\ref{alg:ker}.

\begin{algorithm}
\caption{Knowledgeable Example Retrieval}
\label{alg:ker}
\begin{algorithmic}[1]
\REQUIRE Training set $\mathcal{D}_{trn}$, Target (testing) set $\mathcal{D}_{tgt}$, number of in-context examples $K$.
\STATE Randomly sampling a subset $\mathcal{D}'_{trn}$ from $\mathcal{D}_{trn}$;
\FOR {each target example $(X_j^{tgt})\in\mathcal{D}_{tgt}$}
\STATE Extract entities $E_j^{tgt}$ from this target example;
\FOR {each training example $(X_i^{trn}, y_i^{trn})\in\mathcal{D}'_{trn}$}
\STATE Extract entities $E_i^{trn}$ from this training example;
\STATE Calculate Jaccard similarity $d_{jac}(i, j)$ and Euclidean distance $d_{sem}(i, j)$;
\ENDFOR
\STATE Conditioning on the target example $X_j^{tgt}$, obtain the knowledge relevance score $d(X_i^{trn}, X_j^{tgt})$ for the training example $X_i^{trn}$; 
\ENDFOR
\STATE Calculate the final sampling weight $s'(X_i^{trn})$ for each training example $X_i^{trn}$ in Eq.~\ref{eqn:sample_weight};
\STATE Sampling $K$ training examples via the weight $s'(X_i^{trn})$;
\RETURN The selected $K$ training examples.
\end{algorithmic}
\end{algorithm}

\subsection{Prediction Details}
We first provide the details of the prompt formats and label mapping rules.
Specifically, for the classification task, we need to define a template and label mapping to guide the model to generate results toward pre-defined classes. The prompt formats and label words are shown in Table~\ref{tab:format1}.
For the question answering task, we only need to define the template format, shown in Table~\ref{tab:format2}.

During the prediction, we calibrate the prediction probability. We thus provide the implementation details. 
We obtain a subset of training corpora from the KQA pre-training task, which consists of many question answer pairs. Thus, for each question, we can generate an answer (may be an entity or a label word) at the output position, and obtain the contextualized prior via Eq.~\ref{eqn:context_prior}. The value $P(v)$ means the prior information of the generated entity or label word. Intuitively, if the value $P(v)$ is higher, the entity or label word $v$ is more likely to be generated. We can save these prior values before prediction for downstream tasks.
During the prediction, we can use the prior information of each pre-defined label word or entity to calibrate the prediction probability via Eq.~\ref{eqn:calibrate_prediction}.

\begin{figure*}
\centering
\includegraphics[width=\linewidth]{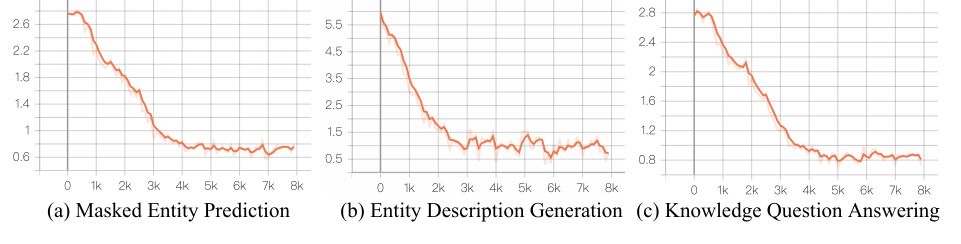}
\vspace{-1.5em}
\caption{The curves of the pre-training loss on GPT-2 (large) for each self-supervised learning task.}
\label{fig:pretrain_curve}
\end{figure*}

\begin{figure}[t]
\centering
\includegraphics[width = 0.7\linewidth]{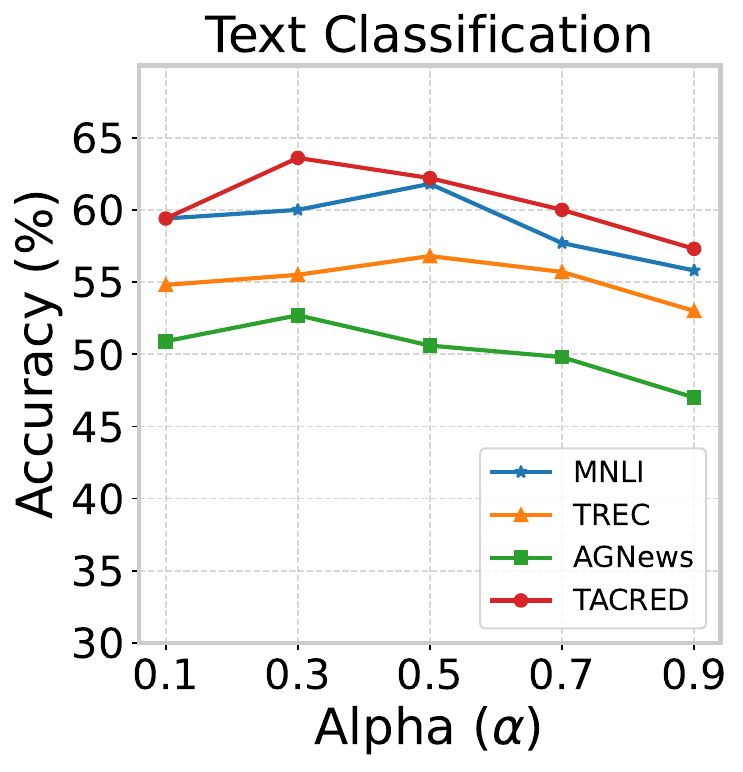}
\includegraphics[width = 0.7\linewidth]{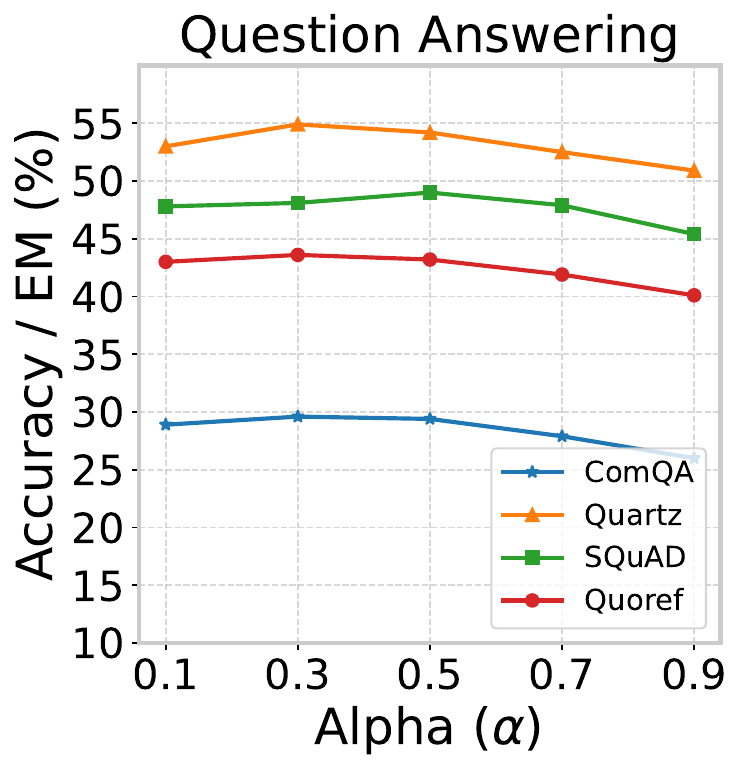}
\caption{The 8-shot performance (\%) of GPT-2 (large) with different $\alpha$ over text classification and question answering tasks.}
\label{fig:alpha}
\end{figure}

\section{Analysis of Settings of Model Variants}
\label{appendix:analysis-settings}
We conduct some detailed analysis of our proposed technique.

\paragraph{Analysis of Pre-training Efficiency.}
To show the efficiency of pre-training, we choose GPT-2 (large) draw the pre-training loss for each self-supervised learning task. 
From Figure~\ref{fig:pretrain_curve},
we can see that as the training process proceeds, each self-supervised learning task has reached the convergence of the model through the entire pre-training process.

\paragraph{Effectiveness of Hyper-parameters.}
In~{\model}, we investigate the effectiveness of the hyper-parameter $\alpha$ in KER, which aims to balance the relevance scores between Jaccard similarity and Euclidean distance.
Results shown in Figure~\ref{fig:alpha} demonstrate that the hyper-parameter $\alpha$ is key to the performance. We can see that the suitable value is around 0.3.

\paragraph{Effectiveness of the Template.}
We believe that the model performances rely on the format of the template, which has been investigated in~\cite{Liu2022What, Min2022Rethinking}.
We choose some other templates for evaluation. For example, when we change the prefix string (e.g., ``Question:'', ``Answer:'') to others (e.g., ``Q:'', ``A:''), the performance improvement of~{\model} is consistent.
In addition, we also find that the text split character `` $\backslash$n'' between each sentence or example is important to support the generation, which is also found in~\cite{Dong2023A, andrew2007scalable, Kim2022Ground, Si2022Prompting}.



\begin{table}
\centering
\resizebox{\linewidth}{!}{
\begin{tabular}{lc}
\toprule
\bf Hyper-parameter &\bf Value \\
\midrule
Batch Size & \{2, 4, 8, 16, 32, 64\} \\
Seed & \{12, 24, 42, 90, 100\} \\
$K$ & \{0, 1, 4, 8, 16\} \\
$\alpha$ & \{0.1, 0.3, 0.5, 0.7, 0.9\} \\
$\gamma$ & \{0.001, 0.01, 0.05, 0.1, 0.5, 1.0\} \\
\bottomrule
\end{tabular}
}
\caption{The searching scope for each hyper-parameter.}
\label{tab:search-scope}
\end{table}

\begin{table*}
\begin{center}
\centering
\centering
\begin{small}
\begin{tabular}{p{1.0cm}p{11.4cm}p{2.0cm}}
\toprule
\textbf{Task} & \textbf{Prompt} & \textbf{Label Words} \\
\midrule
{SST-2} & Review: This movie is amazing! \newline Sentiment: Positive
\vspace{3.9pt}\hspace{-0.1cm}

Review: Horrific movie, don't see it. \newline Sentiment: &  Positive, Negative \\

\midrule
{MRPC} & Whether the two questions are similar?\vspace{3.9pt}\hspace{-0.1cm}

Question 1: How much is this book? Question 2: How many books? \newline Output: No
\vspace{3.9pt}\hspace{-0.1cm}

Question 1: Do you know the reason? Question 2: What's the reason? \newline Output: &  Yes, No \\

\midrule
{MNLI} & Is entailment, neutral, or contradiction between two texts?\vspace{3.9pt}\hspace{-0.1cm}

Text 1: We sought to identify practices within the past 5 years.
Text 2: We want to identify practices commonly used by agencies in the last 5 years. \newline Output: entailment
\vspace{3.9pt}\hspace{-0.1cm}

Text 1: yeah well you're a student right Text 2: Well you're a mechanics student right? Output: &  entailment, neutral, contradiction \\

\midrule
{QNLI} & Whether the answer is entailed to the question?\vspace{3.9pt}\hspace{-0.1cm}

Text 1: In what year did the university first see a drop in applications? Text2: In the early 1950s, student applications declined as a result of increasing crime and $\cdots$ \newline Output: Yes
\vspace{3.9pt}\hspace{-0.1cm}

Text1: When did Tesla move to Gospic? Text2: Tesla was the fourth of five children. \newline Output: &  Yes, No \\

\midrule
{RTE} & Others argue that Mr. Sharon should have negotiated the Gaza pullout - both to obtain at least some written promises of $\cdots$ \newline
Question: Mr. Abbas is a member of the Palestinian family. True or False?\newline
Answer: False
\vspace{3.9pt}\hspace{-0.1cm}

The program will include Falla's "Night in the Gardens of Spain," Ravel's Piano $\cdots$ \newline
Question: Beatrice and Benedict is an overture by Berlioz. True or False?\newline
Answer: & True, False \\

\midrule
CB & But he ended up eating it himself. I was reluctant to kiss my mother, afraid that somehow her weakness and unhappiness would infect me. $\cdots$ \newline
Question: her life and spirit could stimulate her mother. True, False, or Neither?\newline
Answer: Neither
\vspace{3.9pt}\hspace{-0.1cm}

Valence the void-brain, Valence the virtuous valet. Why couldn't the figger choose his own portion of titanic anatomy to shaft? Did he think he was helping?\newline
Question: Valence was helping. True, False, or Neither?\newline
Answer: & True, False, Neither \\

\midrule
{TREC} & Classify the questions based on whether their answer type is a Number, Location, Person, Description, Entity, or Abbreviation. \vspace{3.9pt}\hspace{-0.1cm}

Question: How did serfdom develop in and then leave Russia? \newline
Answer Type: Description
\vspace{3.9pt}\hspace{-0.1cm}

Question: When was Ozzy Osbourne born? \newline
Answer Type: &  Number, Location, Person, Description, Entity, Abbreviation \\

\midrule
{AGNews} & Article: USATODAY.com - Retail sales bounced back a bit in July, and new claims for jobless benefits fell last week, the government said Thursday, indicating $\cdots$ \newline Answer: Business
\vspace{3.9pt}\hspace{-0.1cm}

Article: New hard-drive based devices feature color screens, support for WMP 10. \newline Answer: &  World, Sports, Business, Technology \\





\bottomrule
\end{tabular}
\end{small}
\end{center}
\vspace{-0.2cm}
\caption{The prompts used for text classification. We show one training example per task for illustration purposes. The right column shows the label words (aiming to map the word to the original label class).}
\label{tab:format1}
\end{table*}

\begin{table*}
\begin{center}
\centering
\centering
\begin{small}
\begin{tabular}{p{1.0cm}p{13.4cm}}
\toprule
\textbf{Task} & \textbf{Prompt} \\
\midrule
{ComQA} & Answer the question through multiple-choice.
\vspace{3.9pt}\hspace{-0.1cm}

Question: When people want to watch a new move, the often go see it at the? (A) town (B) conference (C) bathroom (D) theater (E) train station \newline Answer: theater
\vspace{3.9pt}\hspace{-0.1cm}

Question: Where is known to always have snow? (A) africa (B) north pole (C) roof (D) canada (E) surface of earth  north pole \newline Answer: \\

\midrule
{Quartz} & Answer the question through multiple-choice.
\vspace{3.9pt}\hspace{-0.1cm}

Question: Eric pushes an electron closer to the nucleus of an atom. The electron \_\_\_\_\_ energy.As you go farther from the nucleus of an atom, the electron levels have more and more energy. (A) loses (B) gains \newline Answer: gains
\vspace{3.9pt}\hspace{-0.1cm}

Question: When something is very lightweight what does it need to move?Objects with greater mass have greater inertia. (A) more inertia (B) less inertia \newline Answer: \\

\midrule
{SQuAD} & Read the question and find an answer in the context.
\vspace{3.9pt}\hspace{-0.1cm}

Question: Where was the first figure skating championship held? \newline Context: The tourism industry began in the early 19th century when foreigners visited the Alps, traveled to the bases of the mountains to enjoy the scenery, and stayed at the spa-resorts. Large hotels were built during the Belle Époque; cog-railways, built early in the 20th century, brought tourists to ever higher elevations, with the Jungfraubahn terminating at the Jungfraujoch, well above the eternal snow-line, after going through a tunnel in Eiger. During this period winter sports were slowly introduced: in 1882 the first figure skating championship was held in St. Moritz, and downhill skiing became a popular sport with English visitors early in the 20th century, as the first ski-lift was installed in 1908 above Grindelwald. \newline Answer: St. Moritz
\vspace{3.9pt}\hspace{-0.1cm}

Question: What are some examples of classical violinists from Portugal? \newline 
Context: In the classical music domain, Portugal is represented by names as the pianists Artur Pizarro, Maria João Pires, Sequeira Costa, the violinists Carlos Damas, Gerardo Ribeiro and in the past by the great cellist Guilhermina Suggia. Notable composers include José Vianna da Motta, Carlos Seixas, João Domingos Bomtempo, João de Sousa Carvalho, Luís de Freitas Branco and his student Joly Braga Santos, Fernando Lopes-Graça, Emmanuel Nunes and Sérgio Azevedo. Similarly, contemporary composers such as Nuno Malo and Miguel d'Oliveira have achieved some international success writing original music for film and television. \newline Answer: \\

\midrule
{Quoref} & Read the question and find an answer in the context.
\vspace{3.9pt}\hspace{-0.1cm}

Question: What's the name of the person whose birth causes Sarah to die? \newline 
Context: Jack and Sarah are expecting a baby together, but a complication during the birth leads to the death of Sarah. Jack, grief-stricken, goes on an alcoholic bender, leaving his daughter to be taken care of by his parents and Sarah's mother, until they decide to take drastic action: they return the baby to Jack whilst he is asleep, leaving him to take care of it. $\cdots$ \newline Answer: Sarah
\vspace{3.9pt}\hspace{-0.1cm}

Question: What is the first name of the person the actor believes is a little too odd? \newline
Context: When a British secret agent is murdered in the line of duty, agent Karen Bentley inherits the mission from her partner. The mission is to deliver a flight plan for a hundred American bomber planes to a British agent in Chicago. The plans are hidden in a small medallion of a scorpion that Karen wears. $\cdots$ \newline Answer: \\
\bottomrule
\end{tabular}
\end{small}
\end{center}
\vspace{-0.2cm}
\caption{The prompts used for question answering. We show one training example per task for illustration purposes.}
\label{tab:format2}
\end{table*}

\section{Details of the Grid Search}
\label{app:grid-search}
For the downstream task inference, the searching scope of each model hyper-parameter is shown in Table~\ref{tab:search-scope}.

\section{Performance on Different LLMs}
\label{appdix:more_results}
To show that our method is general and can be applied to other similar models,
we choose other scale sizes of GPT-2 and OPT to show the effectiveness of our~{\model}.
More other experiments results are shown from Table~\ref{tab:main-result-cls-gpt2-small} to Table~\ref{tab:main-result-qa-opt-large}.

\begin{table*}
\centering
\resizebox{\linewidth}{!}{
\begin{tabular}{lcccccccccc}
\toprule
\bf \multirow{2}*{\bf Baselines} & \bf  SST-2 & \bf  MRPC & \bf  MNLI & \bf  QNLI & \bf  RTE & \bf CB & \bf TREC & \bf AGNews & \bf \multirow{2}*{\bf Avg.} \\
& acc & f1 & acc & acc & acc & acc & acc & acc & \\
\midrule
\multicolumn{10}{l}{\textit{\textbf{Full Data}}}\\
Fine Tuning (RoBERTa-large) & 95.00 & 91.40 & 89.80 & 93.30 & 80.90 & 90.50 & 97.40 & 94.70 & 91.63 \\
\midrule
\multicolumn{10}{l}{\textit{\textbf{Few-shot Labeled Data (8-shot)}}}\\
ICL~\cite{Brown2020Language} & 
66.58\small{\textpm4.7} & 44.73\small{\textpm2.5} & 49.80\small{\textpm2.9} & 46.33\small{\textpm2.2} & 45.70\small{\textpm3.8} & 36.92\small{\textpm2.3} & 44.38\small{\textpm2.6} & 40.53\small{\textpm4.0} & 46.87 \\
CBU~\cite{Zhao2021Calibrate} & 
74.19\small{\textpm4.1} & 48.88\small{\textpm3.3} & 51.10\small{\textpm2.5} & 48.39\small{\textpm3.2} & 40.07\small{\textpm3.0} & 39.26\small{\textpm2.8} & 47.94\small{\textpm2.2} & 43.28\small{\textpm2.2} & 49.14 \\
KATE~\cite{Liu2022What} & 
72.38\small{\textpm2.9} & 46.38\small{\textpm3.2} & 49.15\small{\textpm3.0} & 47.28\small{\textpm2.8} & 46.30\small{\textpm2.6} & 41.48\small{\textpm2.1} & 47.80\small{\textpm2.2} & 43.83\small{\textpm3.1} & 49.95 \\
MetaICL$^{\dag}$~\cite{Min2022MetaICL} & 
77.20\small{\textpm3.6} & 51.21\small{\textpm2.5} & 53.29\small{\textpm3.0} & 49.42\small{\textpm2.2} & 48.33\small{\textpm2.0} & 40.18\small{\textpm1.9} & 49.68\small{\textpm2.8} & 47.35\small{\textpm2.9} & 52.08 \\
SelfSup.$^{\dag}$~\cite{Chen2022Improving} & 
78.94\small{\textpm3.0} & 52.13\small{\textpm2.0} & 52.70\small{\textpm2.2} & 48.29\small{\textpm1.8} & 49.27\small{\textpm2.6} & 41.80\small{\textpm2.5} & 48.59\small{\textpm2.5} & 47.39\small{\textpm3.2} & 52.39 \\
\hdashline
{\model}$^{\dag}$ & \bf
82.18\small{\textpm3.2} & \bf 54.19\small{\textpm3.7} & \bf 54.85\small{\textpm2.3} & \bf 50.93\small{\textpm1.9} & \bf 50.13\small{\textpm2.2} & \bf 43.89\small{\textpm2.8} & \bf 51.38\small{\textpm2.5} & \bf 51.20\small{\textpm3.0} & \bf 54.90 \\
\bottomrule
\end{tabular}
}
\caption{The 8-shot performance (\%) on GPT-2 (small) of different learning settings with standard deviations over text classification benchmarks. $^{\dag}$ denotes the method involves parameters update for ICL.}
\label{tab:main-result-cls-gpt2-small}
\end{table*}

\begin{table*}
\centering
\resizebox{\linewidth}{!}{
\begin{tabular}{lcccccccccc}
\toprule
\bf \multirow{2}*{\bf Baselines} & \bf  SST-2 & \bf  MRPC & \bf  MNLI & \bf  QNLI & \bf  RTE & \bf CB & \bf TREC & \bf AGNews & \bf \multirow{2}*{\bf Avg.} \\
& acc & f1 & acc & acc & acc & acc & acc & acc & \\
\midrule
\multicolumn{10}{l}{\textit{\textbf{Full Data}}}\\
Fine Tuning (RoBERTa-large) & 95.00 & 91.40 & 89.80 & 93.30 & 80.90 & 90.50 & 97.40 & 94.70 & 91.63 \\
\midrule
\multicolumn{10}{l}{\textit{\textbf{Few-shot Labeled Data (8-shot)}}}\\
ICL~\cite{Brown2020Language} & 
71.39\small{\textpm3.2} & 49.60\small{\textpm2.8} & 53.90\small{\textpm2.4} & 50.04\small{\textpm3.2} & 51.18\small{\textpm4.1} & 39.33\small{\textpm2.8} & 49.20\small{\textpm2.1} & 43.75\small{\textpm3.6} & 51.05 \\
CBU~\cite{Zhao2021Calibrate} & 
77.71\small{\textpm3.8} & 55.48\small{\textpm3.1} & 55.41\small{\textpm2.2} & 51.10\small{\textpm3.0} & 47.53\small{\textpm2.8} & 48.11\small{\textpm2.7} & 51.52\small{\textpm2.7} & 53.27\small{\textpm2.4} & 55.02 \\
KATE~\cite{Liu2022What} & 
75.32\small{\textpm3.1} & 53.80\small{\textpm3.1} & 48.88\small{\textpm3.4} & 50.14\small{\textpm2.5} & 45.82\small{\textpm2.9} & 47.05\small{\textpm2.4} & 50.25\small{\textpm2.8} & 51.93\small{\textpm3.4} & 52.89 \\
MetaICL$^{\dag}$~\cite{Min2022MetaICL} & 
80.16\small{\textpm3.0} & 61.33\small{\textpm2.0} & 56.12\small{\textpm3.1} & 54.24\small{\textpm2.9} & 54.93\small{\textpm2.9} & 46.50\small{\textpm2.9} & 53.22\small{\textpm2.8} & 53.36\small{\textpm2.4} & 57.48 \\
SelfSup.$^{\dag}$~\cite{Chen2022Improving} & 
81.62\small{\textpm3.0} & 58.43\small{\textpm3.2} & 59.53\small{\textpm2.6} & 51.70\small{\textpm3.8} & 54.33\small{\textpm2.6} & 43.48\small{\textpm3.5} & 53.46\small{\textpm2.6} & 53.73\small{\textpm3.1} & 57.04 \\
\hdashline
{\model}$^{\dag}$ & \bf
89.10\small{\textpm3.9} & \bf 66.44\small{\textpm2.7} & \bf 64.85\small{\textpm3.0} & \bf 57.81\small{\textpm3.2} & \bf 61.02\small{\textpm4.0} & \bf 53.91\small{\textpm2.3} & \bf 60.34\small{\textpm2.0} & \bf 61.77\small{\textpm3.3} & \bf 64.41 \\
\bottomrule
\end{tabular}
}
\caption{The 8-shot performance (\%) on GPT-2 (medium) of different learning settings with standard deviations over text classification benchmarks. $^{\dag}$ denotes the method involves parameters update for ICL.}
\label{tab:main-result-cls-gpt2-base}
\end{table*}

\begin{table*}
\centering
\resizebox{\linewidth}{!}{
\begin{tabular}{lcccccccccc}
\toprule
\bf \multirow{2}*{\bf Baselines} & \bf  SST-2 & \bf  MRPC & \bf  MNLI & \bf  QNLI & \bf  RTE & \bf CB & \bf TREC & \bf AGNews & \bf \multirow{2}*{\bf Avg.} \\
& acc & f1 & acc & acc & acc & acc & acc & acc & \\
\midrule
\multicolumn{10}{l}{\textit{\textbf{Full Data}}}\\
Fine Tuning (RoBERTa-large) & 95.00 & 91.40 & 89.80 & 93.30 & 80.90 & 90.50 & 97.40 & 94.70 & 91.63 \\
\midrule
\multicolumn{10}{l}{\textit{\textbf{Few-shot Labeled Data (8-shot)}}}\\
ICL~\cite{Brown2020Language} & 
78.98\small{\textpm7.2} & 56.36\small{\textpm2.3} & 58.25\small{\textpm2.4} & 55.03\small{\textpm3.2} & 55.01\small{\textpm5.0} & 44.04\small{\textpm1.8} & 53.29\small{\textpm4.1} & 47.33\small{\textpm6.6} & 56.04 \\
CBU~\cite{Zhao2021Calibrate} & 
83.31\small{\textpm4.4} & 65.17\small{\textpm3.9} & 58.13\small{\textpm2.8} & 55.59\small{\textpm3.9} & 55.97\small{\textpm2.8} & 53.14\small{\textpm1.7} & 56.29\small{\textpm3.7} & 57.89\small{\textpm2.8} & 60.69 \\
KATE~\cite{Liu2022What} & 
82.55\small{\textpm3.8} & 59.43\small{\textpm3.9} & 61.20\small{\textpm2.4} & 55.37\small{\textpm3.5} & 55.57\small{\textpm2.7} & 48.27\small{\textpm2.1} & 56.11\small{\textpm2.8} & 53.78\small{\textpm3.4} & 59.04 \\
MetaICL$^{\dag}$~\cite{Min2022MetaICL} & 
88.80\small{\textpm5.0} & 64.22\small{\textpm2.0} & 62.39\small{\textpm3.4} & 57.34\small{\textpm1.9} & 59.18\small{\textpm2.8} & 50.46\small{\textpm2.5} & 57.90\small{\textpm1.8} & 57.13\small{\textpm2.4} & 62.18 \\
SelfSup.$^{\dag}$~\cite{Chen2022Improving} & 
88.55\small{\textpm3.0} & 64.24\small{\textpm2.0} & 63.42\small{\textpm2.2} & 55.70\small{\textpm1.8} & 58.93\small{\textpm2.6} & 48.08\small{\textpm2.5} & 58.01\small{\textpm2.5} & 58.28\small{\textpm3.2} & 61.90 \\
\hdashline
{\model}$^{\dag}$ & \bf
92.18\small{\textpm2.9} & \bf 71.32\small{\textpm0.7} & \bf 71.23\small{\textpm1.0} & \bf 62.89\small{\textpm1.2} & \bf 66.10\small{\textpm4.2} & \bf 58.33\small{\textpm3.8} & \bf 64.90\small{\textpm5.5} & \bf 69.27\small{\textpm5.7} & \bf 69.53 \\
\bottomrule
\end{tabular}
}
\caption{The 8-shot performance (\%) on GPT-2 (urge) of different learning settings with standard deviations over text classification benchmarks. $^{\dag}$ denotes the method involves parameters update for ICL.}
\label{tab:main-result-cls-gpt2-urge}
\end{table*}

\begin{table*}
\centering
\resizebox{\linewidth}{!}{
\begin{tabular}{lcccccccccc}
\toprule
\bf \multirow{2}*{\bf Baselines} & \bf  SST-2 & \bf  MRPC & \bf  MNLI & \bf  QNLI & \bf  RTE & \bf CB & \bf TREC & \bf AGNews & \bf \multirow{2}*{\bf Avg.} \\
& acc & f1 & acc & acc & acc & acc & acc & acc & \\
\midrule
\multicolumn{10}{l}{\textit{\textbf{Full Data}}}\\
Fine Tuning (RoBERTa-large) & 95.00 & 91.40 & 89.80 & 93.30 & 80.90 & 90.50 & 97.40 & 94.70 & 91.63 \\
\midrule
\multicolumn{10}{l}{\textit{\textbf{Few-shot Labeled Data (8-shot)}}}\\
ICL~\cite{Brown2020Language} & 
79.43\small{\textpm7.2} & 56.72\small{\textpm2.3} & 59.28\small{\textpm2.4} & 55.37\small{\textpm3.2} & 56.01\small{\textpm5.0} & 44.48\small{\textpm1.8} & 54.10\small{\textpm4.1} & 47.95\small{\textpm6.6} & 56.67 \\
CBU~\cite{Zhao2021Calibrate} & 
83.77\small{\textpm4.4} & 65.38\small{\textpm3.9} & 58.49\small{\textpm2.8} & 55.88\small{\textpm3.9} & 56.26\small{\textpm2.8} & 53.89\small{\textpm1.7} & 56.37\small{\textpm3.7} & 58.20\small{\textpm2.8} & 61.03 \\
KATE~\cite{Liu2022What} & 
83.18\small{\textpm3.8} & 59.83\small{\textpm3.9} & 62.40\small{\textpm2.4} & 55.87\small{\textpm3.5} & 55.81\small{\textpm2.7} & 48.83\small{\textpm2.1} & 56.98\small{\textpm2.8} & 54.32\small{\textpm3.4} & 59.65 \\
MetaICL$^{\dag}$~\cite{Min2022MetaICL} & 
90.03\small{\textpm5.0} & 64.72\small{\textpm2.0} & 62.99\small{\textpm3.4} & 57.94\small{\textpm1.9} & 59.81\small{\textpm2.8} & 51.29\small{\textpm2.5} & 58.50\small{\textpm1.8} & 58.12\small{\textpm2.4} & 62.93 \\
SelfSup.$^{\dag}$~\cite{Chen2022Improving} & 
88.59\small{\textpm3.0} & 64.24\small{\textpm2.0} & 64.42\small{\textpm2.2} & 56.60\small{\textpm1.8} & 59.22\small{\textpm2.6} & 49.58\small{\textpm2.5} & 59.33\small{\textpm2.5} & 59.48\small{\textpm3.2} & 62.77 \\
\hdashline
{\model}$^{\dag}$ & \bf
92.38\small{\textpm2.9} & \bf 71.92\small{\textpm0.7} & \bf 71.83\small{\textpm1.0} & \bf 63.21\small{\textpm1.2} & \bf 66.83\small{\textpm4.2} & \bf 58.70\small{\textpm3.8} & \bf 65.38\small{\textpm5.5} & \bf 70.42\small{\textpm5.7} & \bf 70.08 \\
\bottomrule
\end{tabular}
}
\caption{The 8-shot performance (\%) on OPT (large) of different learning settings with standard deviations over text classification benchmarks. $^{\dag}$ denotes the method involves parameters update for ICL.}
\label{tab:main-result-cls-opt-large}
\end{table*}

\begin{table}[h]
\centering
\resizebox{\linewidth}{!}{
\begin{tabular}{lcccccc}
\toprule
\bf \multirow{2}*{\bf Baselines} & \bf  ComQA & \bf  Quartz & \bf  SQuAD & \bf  Quoref & \bf \multirow{2}*{\bf Avg.} \\
& acc & acc & em & em & \\
\midrule
\multicolumn{6}{l}{\textit{\textbf{Full Data}}}\\
Fine Tuning (RoBERTa-large) & 72.10 & 76.90 & 86.50 & 78.70 & 78.55 \\
\midrule
\multicolumn{6}{l}{\textit{\textbf{Few Labeled Data (8-shot)}}}\\
ICL~\cite{Brown2020Language} & 
23.70\small{\textpm3.7} & 49.20\small{\textpm1.9} & 43.10\small{\textpm3.4} & 37.30\small{\textpm3.0} & 38.34 \\
CBU~\cite{Zhao2021Calibrate} & 
26.37\small{\textpm3.1} & 52.90\small{\textpm2.8} & 46.88\small{\textpm2.0} & 41.38\small{\textpm2.9} & 41.89 \\
KATE~\cite{Liu2022What} & 
26.89\small{\textpm3.2} & 52.88\small{\textpm3.1} & 46.93\small{\textpm3.7} & 41.35\small{\textpm2.8} & 42.01 \\
MetaICL$^{\dag}$~\cite{Min2022MetaICL} & 
27.40\small{\textpm2.7} & 52.74\small{\textpm3.3} & 46.63\small{\textpm2.9} & 42.51\small{\textpm3.0} & 42.32 \\
SelfSup.$^{\dag}$~\cite{Chen2022Improving} & 
27.33\small{\textpm3.1} & 52.91\small{\textpm3.1} & 46.97\small{\textpm2.9} & 42.71\small{\textpm3.2} & 42.48 \\
\hdashline
{\model}$^{\dag}$ & 
\bf 28.78\small{\textpm2.6} & \bf 53.10\small{\textpm2.9} & \bf 47.72\small{\textpm2.3} & \bf 43.88\small{\textpm2.2} & \bf 43.37 \\
\bottomrule
\end{tabular}
}
\caption{The 8-shot performance (\%) on GPT-2 (small) of different learning settings with standard deviations over question answering benchmarks.}
\label{tab:main-result-qa-gpt2-small}
\end{table}

\begin{table}
\centering
\resizebox{\linewidth}{!}{
\begin{tabular}{lcccccc}
\toprule
\bf \multirow{2}*{\bf Baselines} & \bf  ComQA & \bf  Quartz & \bf  SQuAD & \bf  Quoref & \bf \multirow{2}*{\bf Avg.} \\
& acc & acc & em & em & \\
\midrule
\multicolumn{6}{l}{\textit{\textbf{Full Data}}}\\
Fine Tuning (RoBERTa-large) & 72.10 & 76.90 & 86.50 & 78.70 & 78.55 \\
\midrule
\multicolumn{6}{l}{\textit{\textbf{Few Labeled Data (8-shot)}}}\\
ICL~\cite{Brown2020Language} & 
25.38\small{\textpm3.1} & 52.10\small{\textpm3.2} & 45.58\small{\textpm3.3} & 38.47\small{\textpm2.7} & 40.38 \\
CBU~\cite{Zhao2021Calibrate} & 
28.40\small{\textpm3.2} & 53.64\small{\textpm2.6} & 47.81\small{\textpm4.0} & 43.20\small{\textpm2.2} & 42.68 \\
KATE~\cite{Liu2022What} & 
28.38\small{\textpm3.1} & 54.26\small{\textpm3.3} & 46.70\small{\textpm3.7} & 41.98\small{\textpm4.1} & 42.83 \\
MetaICL$^{\dag}$~\cite{Min2022MetaICL} & 
29.67\small{\textpm2.9} & 54.37\small{\textpm2.5} & 48.79\small{\textpm2.4} & 45.11\small{\textpm3.1} & 44.49 \\
SelfSup.$^{\dag}$~\cite{Chen2022Improving} & 
29.36\small{\textpm3.0} & 54.10\small{\textpm2.2} & 48.47\small{\textpm2.7} & 44.06\small{\textpm3.1} & 44.00 \\
\hdashline
{\model}$^{\dag}$ & 
\bf 34.81\small{\textpm3.0} & \bf 56.38\small{\textpm2.9} & \bf 51.18\small{\textpm2.8} & \bf 46.00\small{\textpm3.5} & \bf 47.09 \\
\bottomrule
\end{tabular}
}
\caption{The 8-shot performance (\%) on GPT-2 (medium) of different learning settings with standard deviations over question answering benchmarks.}
\label{tab:main-result-qa-gpt2-base}
\end{table}

\begin{table}
\centering
\resizebox{\linewidth}{!}{
\begin{tabular}{lcccccc}
\toprule
\bf \multirow{2}*{\bf Baselines} & \bf  ComQA & \bf  Quartz & \bf  SQuAD & \bf  Quoref & \bf \multirow{2}*{\bf Avg.} \\
& acc & acc & em & em & \\
\midrule
\multicolumn{6}{l}{\textit{\textbf{Full Data}}}\\
Fine Tuning (RoBERTa-large) & 72.10 & 76.90 & 86.50 & 78.70 & 78.55 \\
\midrule
\multicolumn{6}{l}{\textit{\textbf{Few Labeled Data (8-shot)}}}\\
ICL~\cite{Brown2020Language} & 
29.15\small{\textpm2.4} & 55.78\small{\textpm3.1} & 49.12\small{\textpm3.1} & 42.11\small{\textpm2.7} & 44.04 \\
CBU~\cite{Zhao2021Calibrate} & 
31.58\small{\textpm3.9} & 57.01\small{\textpm2.6} & 51.28\small{\textpm2.8} & 45.70\small{\textpm4.4} & 46.39 \\
KATE~\cite{Liu2022What} & 
31.18\small{\textpm4.1} & 56.70\small{\textpm3.0} & 49.13\small{\textpm3.4} & 44.54\small{\textpm3.3} & 45.39 \\
MetaICL$^{\dag}$~\cite{Min2022MetaICL} & 
32.16\small{\textpm3.2} & 57.64\small{\textpm2.6} & 53.26\small{\textpm3.1} & 48.91\small{\textpm2.9} & 47.99 \\
SelfSup.$^{\dag}$~\cite{Chen2022Improving} & 
33.44\small{\textpm3.2} & 56.18\small{\textpm3.5} & 51.90\small{\textpm2.7} & 49.10\small{\textpm3.1} & 47.66 \\
\hdashline
{\model}$^{\dag}$ & 
\bf 37.05\small{\textpm2.8} & \bf 59.35\small{\textpm2.4} & \bf 55.08\small{\textpm2.9} & \bf 53.18\small{\textpm3.2} & \bf 51.17 \\
\bottomrule
\end{tabular}
}
\caption{The 8-shot performance (\%) on GPT-2 (urge) of different learning settings with standard deviations over question answering benchmarks.}
\label{tab:main-result-qa-gpt2-urge}
\end{table}

\begin{table}
\centering
\resizebox{\linewidth}{!}{
\begin{tabular}{lcccccc}
\toprule
\bf \multirow{2}*{\bf Baselines} & \bf  ComQA & \bf  Quartz & \bf  SQuAD & \bf  Quoref & \bf \multirow{2}*{\bf Avg.} \\
& acc & acc & em & em & \\
\midrule
\multicolumn{6}{l}{\textit{\textbf{Full Data}}}\\
Fine Tuning (RoBERTa-large) & 72.10 & 76.90 & 86.50 & 78.70 & 78.55 \\
\midrule
\multicolumn{6}{l}{\textit{\textbf{Few Labeled Data (8-shot)}}}\\
ICL~\cite{Brown2020Language} & 
30.42\small{\textpm2.2} & 56.19\small{\textpm3.2} & 48.73\small{\textpm3.0} & 44.18\small{\textpm3.7} &  44.88\\
CBU~\cite{Zhao2021Calibrate} & 
32.16\small{\textpm2.7} & 58.02\small{\textpm2.8} & 53.11\small{\textpm2.7} & 47.35\small{\textpm2.0} & 47.66 \\
KATE~\cite{Liu2022What} & 
33.32\small{\textpm3.6} & 58.90\small{\textpm2.9} & 50.65\small{\textpm2.4} & 46.12\small{\textpm3.5} & 47.25 \\
MetaICL$^{\dag}$~\cite{Min2022MetaICL} & 
33.96\small{\textpm3.4} & 58.64\small{\textpm2.4} & 54.11\small{\textpm2.4} & 48.12\small{\textpm2.7} & 48.71 \\
SelfSup.$^{\dag}$~\cite{Chen2022Improving} & 
34.42\small{\textpm3.0} & 58.12\small{\textpm3.0} & 54.92\small{\textpm2.7} & 49.53\small{\textpm1.8} & 49.25 \\
\hdashline
{\model}$^{\dag}$ & 
\bf 39.22\small{\textpm2.8} & \bf 61.71\small{\textpm2.4} & \bf 59.67\small{\textpm2.1} & \bf 54.40\small{\textpm3.1} & \bf 53.75 \\
\bottomrule
\end{tabular}
}
\caption{The 8-shot performance (\%) on OPT (large) of different learning settings with standard deviations over question answering benchmarks.}
\label{tab:main-result-qa-opt-large}
\end{table}

\end{document}